\documentclass{article}

\usepackage{arxiv}

\usepackage[utf8]{inputenc} 
\usepackage[T1]{fontenc}    
\usepackage{hyperref}       
\usepackage{url}            
\usepackage{booktabs}       
\usepackage{amsfonts}       
\usepackage{nicefrac}       
\usepackage{microtype}      
\usepackage{lipsum}		
\usepackage{graphicx}
\usepackage{natbib}
\usepackage{doi}
\usepackage{amsmath}
\usepackage{fancyhdr}       
\usepackage{wrapfig}

\usepackage{adjustbox}
\usepackage{subcaption}
\usepackage{xcolor}

\usepackage{multirow}
\usepackage{makecell} 

\hypersetup{colorlinks,breaklinks,
            citecolor=[RGB]{12,31,105},
            urlcolor=[RGB]{12,31,105},
            linkcolor=[RGB]{12,31,105}}

\title{Direct Training Needs Regularisation: 
            \\ Anytime Optimal Inference Spiking Neural Network
}

\author{Dengyu Wu$^1$, Yi Qi$^1$, Kaiwen Cai$^1$, Gaojie Jin$^2$, Xinping Yi$^3$, Xiaowei Huang$^1$\\
University of Liverpool, Liverpool, UK$^1$\\
State Key Laboratory of Computer Science, Institute of Software, CAS, Beijing, China$^2$\\
Southeast University, Nanjing, China$^3$\\
{\tt\small \{dengyu.wu, xiaowei.huang\}@liverpool.ac.uk}
}

\renewcommand{\shorttitle}{Anytime Optimal Inference SNN}

\hypersetup{
pdftitle={A template for the arxiv style},
pdfsubject={q-bio.NC, q-bio.QM},
pdfauthor={David S.~Hippocampus, Elias D.~Striatum},
pdfkeywords={First keyword, Second keyword, More},
}

\begin{document}
\maketitle

\begin{abstract}
Spiking Neural Network (SNN) is acknowledged as the next generation of Artificial Neural Network (ANN) and hold great promise in effectively processing spatial-temporal information. 
However, the choice of timestep becomes crucial as it significantly impacts the accuracy of the neural network training. 
Specifically, a smaller timestep indicates better performance in efficient computing, resulting in reduced latency and operations. While, using a small timestep may lead to low
accuracy due to insufficient information presentation with few spikes. 
This observation motivates us to develop an SNN that is more reliable for adaptive timestep by introducing a novel regularisation technique, namely Spatial-Temporal Regulariser (STR). Our approach regulates the ratio between the strength of spikes and membrane potential at each timestep. This effectively balances spatial and temporal performance during training, ultimately resulting in an Anytime Optimal Inference (AOI) SNN.
Through extensive experiments on frame-based and event-based datasets, our method, in combination with cutoff based on softmax output,
achieves state-of-the-art performance in terms of both latency and accuracy. 
Notably, with STR and cutoff, SNN achieves $2.14$ to $2.89$ faster in inference compared to the pre-configured 
timestep with near-zero accuracy drop of $0.50\%$ to $0.64\%$ over the event-based datasets. Code available: https://github.com/Dengyu-Wu/AOI-SNN-Regularisation 

\end{abstract}

\section{Introduction}


Spiking Neural Network (SNN) aims to mimic the behavior of biological neurons in the brain, efficiently processing spatial-temporal information through the use of their inherent dynamics, such as integration and firing progress \cite{maass1997networks, rueckauer2017conversion, pfeiffer2018deep, wu2022little}.
For instance, the integrated membrane potential of SNN retains information from previous timestep and enables effective processing of temporal information \cite{yao2021temporal,yin2021accurate}. 
Similarly, the generated spikes activate post neurons, allowing them to efficiently propagate the current information through the network, where neurons are triggered sparsely upon receiving spikes.
This activation mechanism differs from Artificial Neural Network (ANN) that relies on dense multiplications for forward propagation. In SNN, neurons are only activated when receiving spikes which leads to sparse and remarkably efficient computations. 
Given this unique characteristic of SNN, they are particularly well-suited for implementation on emerging neuromorphic hardware platforms,
such as TrueNorth \cite{akopyan2015truenorth}, Loihi \cite{davies2018loihi}, and Tianji \cite{pei2019towards}, which have empowered SNN to leverage their inherent event-driven nature at the hardware level. This development holds great promise for enabling energy-efficient applications, such as real-time audio denoising \cite{timcheck2023intel}, low-power gesture recognition \cite{amir2017low} and robotic control \cite{tang2021deep}.

The rapid progress in SNN has been fueled by the pursuit of energy-efficient and high-performance computing solutions. In the field of neuromorphic computing, the primary focus of algorithm optimisation for SNN has been on improving accuracy \cite{datta2022ace, KULKARNI2018118, 8685186}. From the perspective of the characteristics of SNN, the total inference timestep determines their computing efficiency \cite{rueckauer2017conversion, wu2022little}. Thus, efforts to reduce the inference complexity of SNN while maintaining accuracy have been ongoing. 
Techniques such as optimising SNN through training \cite{deng2022temporal, duan2022temporal, bu2022optimal} have shown promising results in enhancing the computing efficiency of SNN. 
Despite these advancements, there is still room for exploring a more adaptive and flexible inference process, as the current method primarily focuses on optimising SNN for pre-configured timestep. Lately, the concept of anytime inference for SNN has garnered increasing attention, as evidenced by recent works \cite{wu2023optimising, li2023seenn, chen2023spikecp, li2023unleashing}. This growing interest highlights a novel direction of efficient computing for SNN. 
In the meanwhile, \cite{wu2023optimising} suggested that optimising SNN through both training and inference aspects helps achieve Anytime Optimal Inference (AOI). Precisely, a regularisation technique was introduced to train SNN to be \textit{optimal} for anytime inference. 

However, during SNN training, there emerges a delicate balance between optimising the current timestep and considering its potential impact on subsequent ones. As SNN predictions are interconnected, concentrating on minimising loss at one timestep might inadvertently lead to increased loss at the others. This balance highlights the intricate nature of optimising SNN for anytime inference. Simply optimising the average output \cite{wu2019direct,fang2021deep, fang2021incorporating} no longer suffices for achieving anytime optimality, as it lacks constraints at each timestep. While the approach of temporal efficient training \cite{deng2022temporal} comes close to an AOI model by aligning predictions at each timestep closer to the ground truth, it still relies on the average loss across timestep and grapples with the challenge of harmonising the trade-off between spatial and temporal performance.

In this paper, we are interested in optimising SNN for anytime inference through direct training. To achieve this, we introduce a novel regularisation technique that diminishes the influence of present timestep to next timestep, thereby yielding SNN capable of providing more reliable predictions across the timesteps. Our key contributions include: 

\begin{itemize}
\item 
Introducing the concept of the spatial-temporal factor that helps understand the contributions of spatial and temporal information in SNN.
\item Proposing a regularisation technique that dynamically adjusts the spatial-temporal factor during training for enhancing the accuracy at present timestep.
\item Validating our approach with extensive experiments, including uncertainty estimation and cutoff results.
\end{itemize}

Through these contributions, we aim to build a more efficient and accurate SNN for anytime inference. This entails achieving a lower average timestep for the SNN while concurrently maintaining a high level of accuracy.

\section{Related work}
Recent research has extensively explored to reduce the inherent complexity associated with inference processes of SNN. A significant focus within this research landscape is to train SNN that operates at small timestep \cite{deng2021optimal, li2021differentiable, bu2022optimal, duan2022temporal}. Another growing avenue involves the study of adaptive timesteps, providing an alternative to reducing computing operations by lowering average timesteps \cite{wu2023optimising, li2023unleashing, chen2023spikecp}. Both paths exploit the sparsity and dynamics of SNN to achieve efficient computing.

\paragraph{Spiking Network Training} One such direction involves optimising the training of SNN to achieve better efficiency. For instance, reducing the timestep during inference can significantly improve computing efficiency, as the total timestep determine the overall computational operations. This has been achieved by adding temporal batch normalisation \cite{zheng2021going, duan2022temporal}, improving surrogate gradient  \cite{wu2018spatio, wu2019direct, neftci2019surrogate, li2021differentiable}, optimising loss function for temporal training \cite{deng2022temporal}, and minimising the distance between ANN and SNN activation for ANN-to-SNN conversion \cite{deng2021optimal, wu2022little, bu2022optimal}. Another line of investigation focuses on exploring the efficient architecture for SNN, such as designing novel spike-based architectures \cite{fang2021deep, zhou2023spikformer} and deploying Network Architecture Search (NAS) in SNN \cite{kim2022neural, kim2022exploring}. In addition, quantisation techniques \cite{schaefer2020quantizing, putra2021q, li2022quantization}, which aim to convert resource-intensive floating-point operations into more efficient integer operations, have also been explored to enhance the efficiency of SNN. Furthermore, \cite{lu2020exploring} argues that SNN can further benefit from sequential and binarised activation to improve binary network accuracy.  

\paragraph{Anytime Optimal Inference} 
In the realm of SNN, the exploration of anytime inference is still in its early stages, with relatively limited attention garnered thus far. Nonetheless, a few notable studies have begun to delve into enhancing anytime inference in SNN. For example, \cite{li2023seenn} introduced an auxiliary network to predict confidence for the early exiting. Similarly, \cite{chen2023spikecp} studied output distribution and integrated conformal prediction \cite{angelopoulos2021gentle} for adaptive inference. For conversion-based SNN, \cite{li2023unleashing} calibrated output confidence across the timesteps, while \cite{wu2023optimising} suggest that gap value between the first and second largest of outputs can efficiently predict the cutoff time. While these studies efficiently trigger anytime inference in SNN, they have focused on addressing data uncertainty over the inference rather than optimising uncertainty within the SNN model itself. 

\section{Preliminary}
In this section, we introduce the neuron model and direct training of SNN. To facilitate the analysis, we use \textbf{bold~symbol} to represent vector,  $l$ to denote the layer index, and $i$ to denote the index of elements. For example, $\boldsymbol{W}^l$ is weight matrix at the $l$-th layer. $t$ denote discrete timestep. 

\subsection{Leaky Integrate-and-Fire model}
\begin{figure}[t!]
    \centering
    \includegraphics[width=0.68\linewidth]{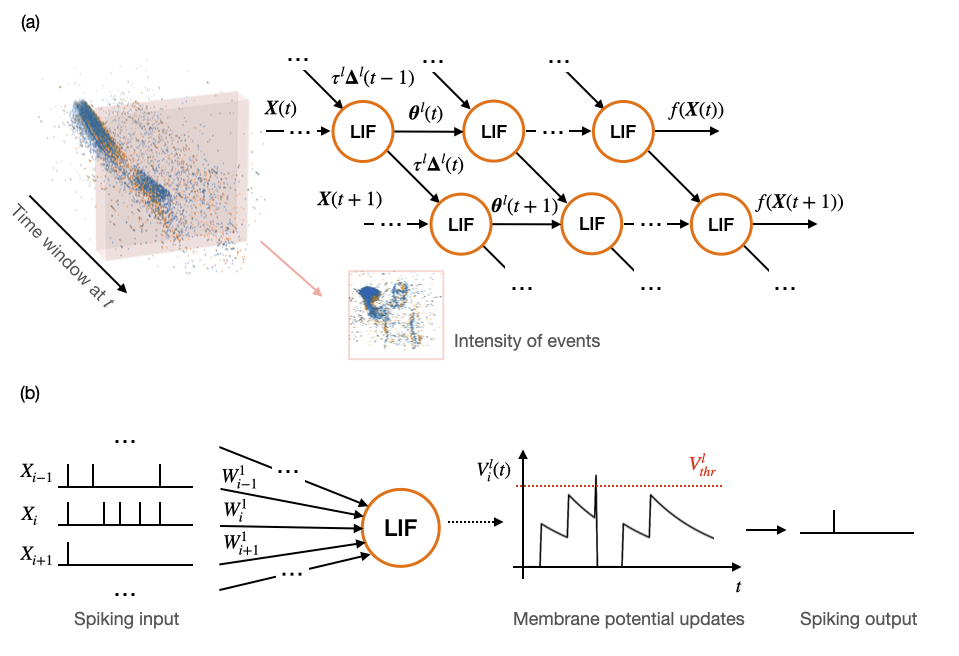}
    \caption{(a) Forward propagation in SNN. The input events $\boldsymbol{X}(t)$ stimulate neurons to generate spikes over the time. The output $f(\boldsymbol{X}(t))$ can respond when a sufficient number of events are received within a specific time window. $\boldsymbol{\theta}(t)$ and $\tau\boldsymbol{\Delta}(t)$ represents the spatial and temporal information at $t$, respectively. 
    (b) The state update of one LIF neuron at input layer during forward propagation process. The weight $W_i^l$ influences the current contributing to the membrane potential $V_i^l(t)$. The threshold $V^l_{thr}$ determines the threshold level of $V_i^l(t)$ required to generate the spikes. 
    }
    \label{fig:snnfp}
\end{figure}

Leaky integrate-and-fire (LIF) model is widely adopted in the study of SNN, due to its simplicity and biological plausibility. The forward propagation in SNN is shown in Figure \ref{fig:snnfp}. The iterative equation of LIF model in forward propagation can be expressed as follows:
\begin{equation} \label{eq:lif}
    \boldsymbol{V}^l(t) = \tau^l\boldsymbol{\Delta}^l(t-1) + \boldsymbol{Z}^l(t), 
\end{equation}
where $\boldsymbol{V}^l(t)$ represents the membrane potential at layer $l$ and time step $t$ prior spike firing, $\tau^l$ denotes decay factor, and $\boldsymbol{\Delta}^l(t) = (1-\boldsymbol{\theta}(t))\cdot \boldsymbol{V}^l(t)$ is the residual current after spike firing, i.e.,  $\theta_i^l(t) = 1 $ if $ V_i^l(t) \geq V^l_{thr}$ and $\theta_i^l (t)= 0 $ otherwise. Furthermore, $\boldsymbol{Z}^l(t)$ denotes current input and is defined as:


\begin{align} \label{eq:lif_init}
    \boldsymbol{Z}^l(t) = \boldsymbol{W}^l\boldsymbol{\theta}^{l-1}(t) + \boldsymbol{b}^l \hspace{0.5cm}\text{ when } l>1.
\end{align}

According to different inputs, $\boldsymbol{Z}^l(t)$ at the first layer, i.e., $\boldsymbol{Z}^1(t)$, can be initialised as:

\begin{equation} \label{eq:lif_input}
    \boldsymbol{Z}^1_t  = 
    \left\{
    \begin{array}{ll}
        \boldsymbol{W}^1\boldsymbol{X}(t) + \boldsymbol{b}^1 &  \textit{event-based input}   \\
        \boldsymbol{W}^1\boldsymbol{\bar X} + \boldsymbol{b}^1 &  \textit{frame-based input}, 
    \end{array}
    \right.
\end{equation}
where $\boldsymbol{X}(t)$ is the integration of events at $t$-th timestep and $\boldsymbol{\bar X}$ represents the constant current stimulus to the first layer that equals to the analogue values of input. Note that for frame-based input, $\boldsymbol{Z}^1(t)$ is the same at different $t$. To simplify the analysis, we simply use 
event-based input as our objective in the following sections. 

\subsection{Direct Training}


Our approach leverages the backpropagation method to train SNN directly. This strategy considers the states of spiking neuron at each timestep during training and has demonstrated the potential to achieve high-performance SNN, particularly when operating with small timesteps. A prevalent optimisation objective aims to minimise the distance between the average output and ground truth, as explored in prior works \cite{fang2021deep, fang2021incorporating, duan2022temporal}. The loss function is defined as:

\begin{equation} \label{eq:mean}
L_{mean} =  L_{ce}(\frac{1}{T} \sum_t^T f(\boldsymbol{X}(t)),\boldsymbol{y}),
\end{equation}
where $T$ is the maximum timestep, $L_{ce}$ represents cross entropy loss, $f(\cdot)$ is the SNN model, and $f(\boldsymbol{X}(t))$ denotes the synaptic current output at $t$-th timestep, and $ \boldsymbol{y} $ is the ground truth. However, recent work \cite{deng2022temporal} proposes an alternative approach, Temporal Efficient Training (TET), to train SNN. TET suggests that employing the average of cross entropy over all timesteps can lead to improved SNN performance, described as:

\begin{equation} \label{eq:tet}
L_{TET} = \frac{1}{T} \sum_t^T L_{ce}(f(\boldsymbol{X}(t)),\boldsymbol{y}).
\end{equation}

Since the firing progress is non-differentiable, surrogate gradient methods, such as linear type \cite{esser2015backpropagation, wu2018spatio, wu2019direct} and non-linear type  \cite{zenke2018superspike, li2021differentiable, shrestha2018slayer}, are employed in direct training. 

\begin{wrapfigure}{r}{0.46\textwidth}
    \begin{center}
    \includegraphics[width=\linewidth]{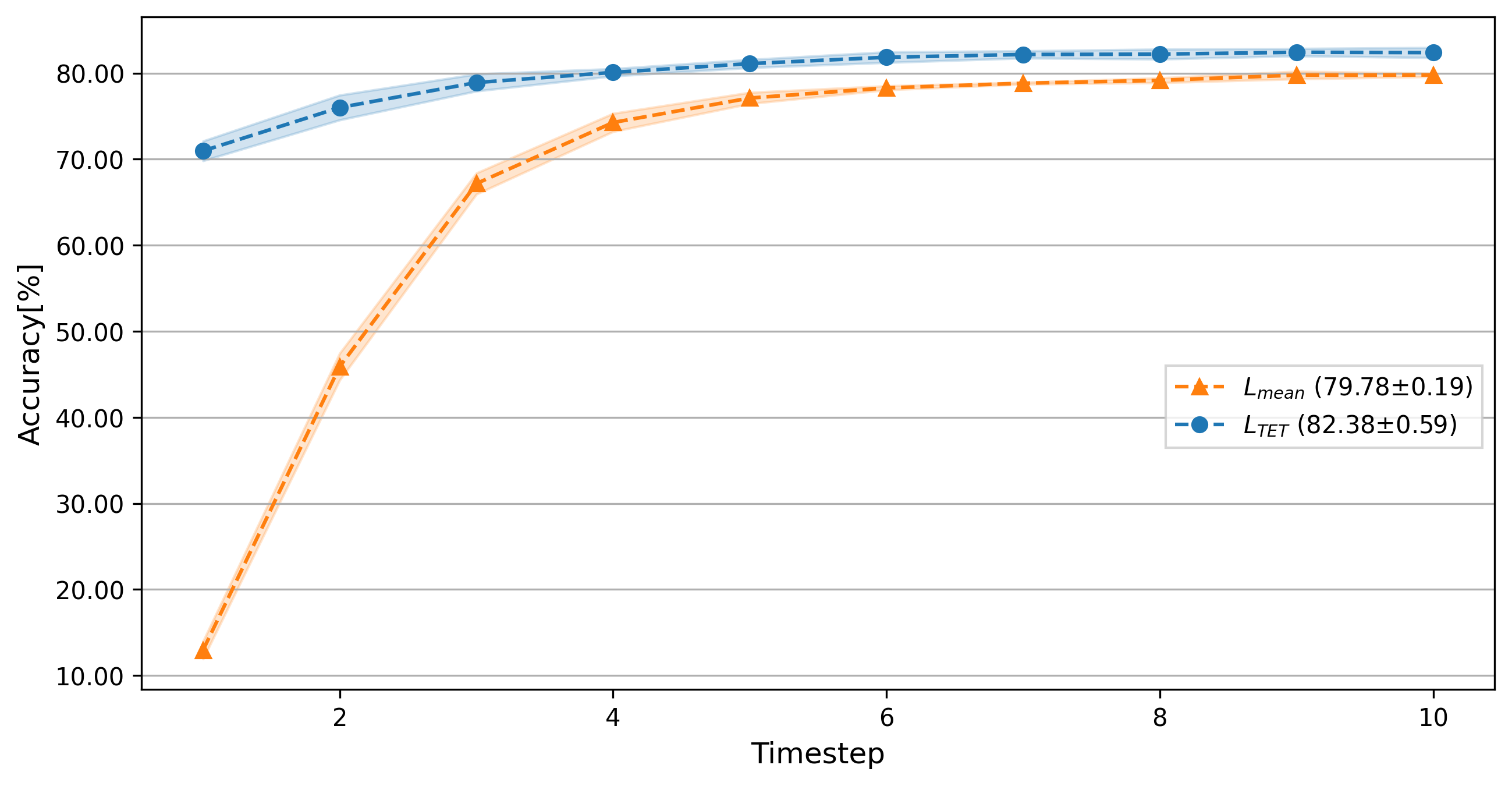}
    \caption{Comparison of accuracy with respect to timestep using different loss functions on Cifar10-DVS.}
    \label{fig:aoi_methods}
    \end{center}
   
\end{wrapfigure}

\paragraph{Training Methods and AOI} 
Figure \ref{fig:aoi_methods} presents accuracy results across all timesteps using two loss functions, e.g., $L_{mean}$ and $L_{TET}$, on Cifar10-DVS \cite{li2017cifar10dvs}. The training of these two models follows the same strategy specified in Section \ref{app:experiment}. It is not surprising that $L_{mean}$ exhibits limited capability in achieving AOI during inference. For example, the accuracy of $L_{mean}$ experiences a significant drop when the timestep is small. This type of training does not train each timestep to yield accurate predictions; instead, it prioritises minimising the loss based on the average output, typically computed at the last timestep. Acknowledging the reliability of TET in achieving AOI, we regularise SNN training through TET in our study.

\section{Method}\label{sec:method}

In this section, we present the details of the two-fold approach that helps SNN to achieve AOI. Firstly, we introduce the Spatial-Temporal Factor (STF) to help better understand how SNN utilise information over different timesteps during inference. Secondly, we propose the Spatial-Temporal Regulariser (STR) to encourage SNN to prioritise the present timestep rather than relying solely on the next timestep to achieve minimal loss during training. 

\subsection{Spatial-Temporal Factor}
To gain a deeper understanding of the forward propagation process of SNN, we decompose the vector $\boldsymbol{V}^l(t)$ into two 
orthogonal components. 
This decomposition facilitates a detailed analysis of the individual contributions made by these components, unveiling the underlying mechanisms involved in information processing within the network. By dissecting the vector $\boldsymbol{V}^l(t)$ in this manner, we can explore how each component influences the dynamics and transformations of information in the SNN. Mathematically, the decomposition is formulated as:
\begin{equation} \label{eq:decompose}
 \boldsymbol{V}^l(t) = V_{thr}^l\boldsymbol{\theta}^l(t) +\tau \boldsymbol{\Delta}^l(t),
\end{equation}
where the clipped value of $\boldsymbol{V}^l(t)$ is ignored as it does not contribute to either the current or the next timestep. 

\begin{figure}[!ht]
    \centering
    \includegraphics[width=0.75\linewidth]{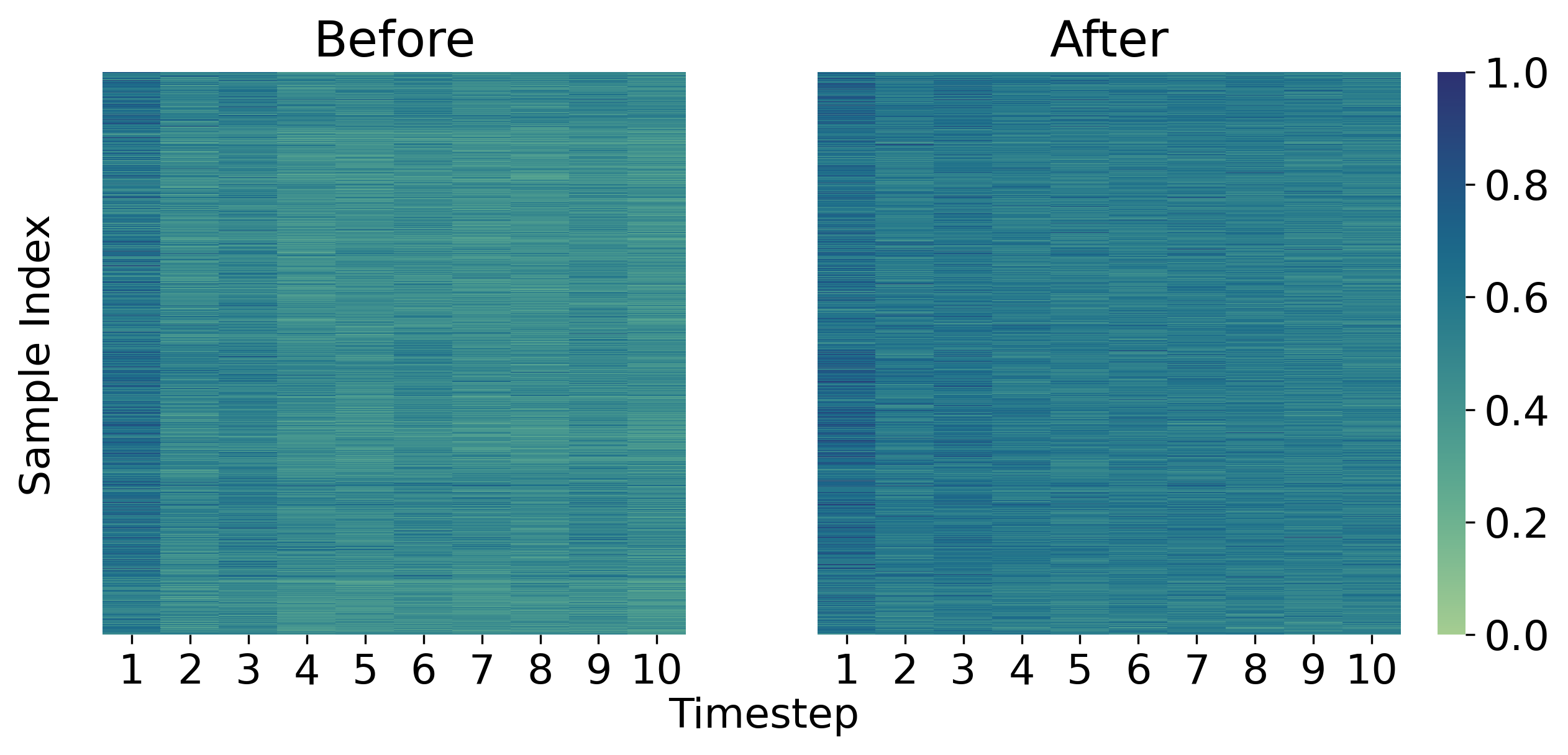}
    \caption{Visualisation of STF $\xi^l(t)$ at 8-th layer in two models, before and after regularisation, on Cifar10-DVS. Regularisation leads to a reduction in variance of $\xi^l(t)$ from 0.0025 to 0.0022, indicating enhanced stability across timestep. Additionally, the mean value of $\xi^l(t)$ rises from 0.2736 to 0.3336, reflecting an enhancement in the representation of spatial information $\lVert\theta^l(t)\rVert_2$.}
    \label{fig:stf_viz}
\end{figure}

Building on this decomposition, we introduce the Spatial-Temporal Factor (STF), symbolised as $\xi^l(t)$, to assess the interplay of $V_{thr}^l\boldsymbol{V}^l(t)$ and $\tau\boldsymbol{\Delta}^l(t)$ on the network. Specifically, the STF is defined as the ratio between the L2 norm of these two elements, offering an approximate estimation of their impacts on the current or future states of the network. This is encapsulated in the following expression:

\begin{equation} \label{eq:reg}
\xi^l(t) 
= \tilde{\alpha} \frac{\lVert \boldsymbol{\theta}^l(t) \rVert_2}{\lVert \boldsymbol{\Delta}^l(t) \rVert_2},
\end{equation}
where $\lVert \cdot \rVert_2$ is the L2 norm, i.e., $\lVert \boldsymbol{x} \rVert_2 = \sqrt{\sum_i x_i^2}$. We define $\tilde \alpha$ as a consolidation of the constants $V_{thr}^l$ and $\tau$, which streamlines the equation and focuses attention on the relationship between $\lVert \boldsymbol{\theta}^l(t) \rVert_2$ and $\lVert \boldsymbol{\Delta}^l(t) \rVert_2$. Given that $\tilde{\alpha}$ is a constant, it can be seamlessly integrated into the hyper-parameter $\alpha$ in Equation \ref{eq:multiloss}. This formulation allows us to focus on the optimisation objective that encompasses both spatial and temporal information within a single term. The influence of regularisation on Equation \ref{eq:reg} is visually demonstrated in Figure \ref{fig:stf_viz}, with a detailed exploration of the regularisation technique presented in Section \ref{sec:reg}. 


\subsection{Spatial-Temporal Regularisation} \label{sec:reg}
\begin{figure}[t!]
    \centering
    \begin{subfigure}[t]{0.48\linewidth}
    \includegraphics[width=0.45\linewidth]{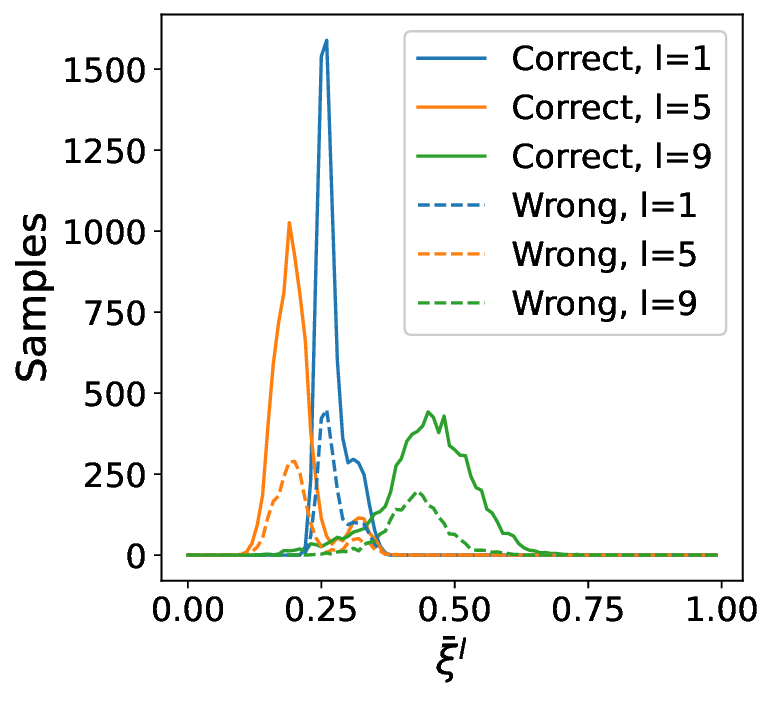}
    \includegraphics[width=0.5\linewidth]{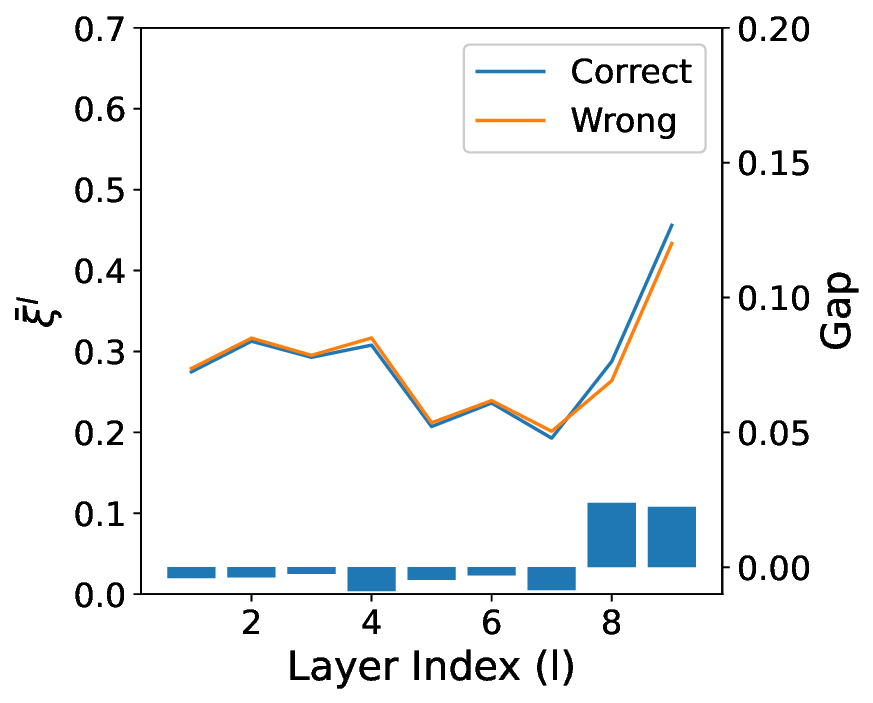}
    \caption{Baseline STF}
    \end{subfigure}
    \begin{subfigure}[t]{0.48\linewidth}
    \includegraphics[width=0.45\linewidth]{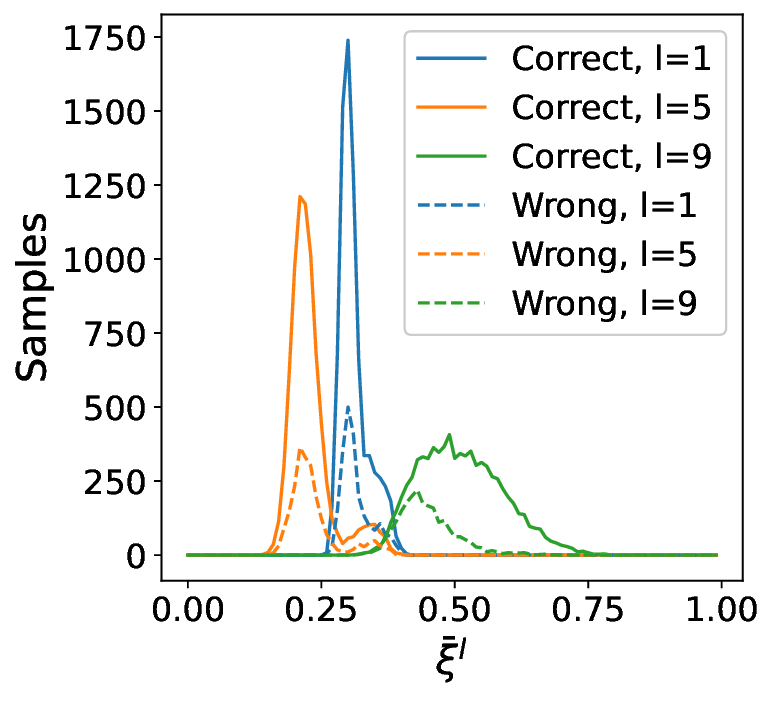}
    \includegraphics[width=0.5\linewidth]{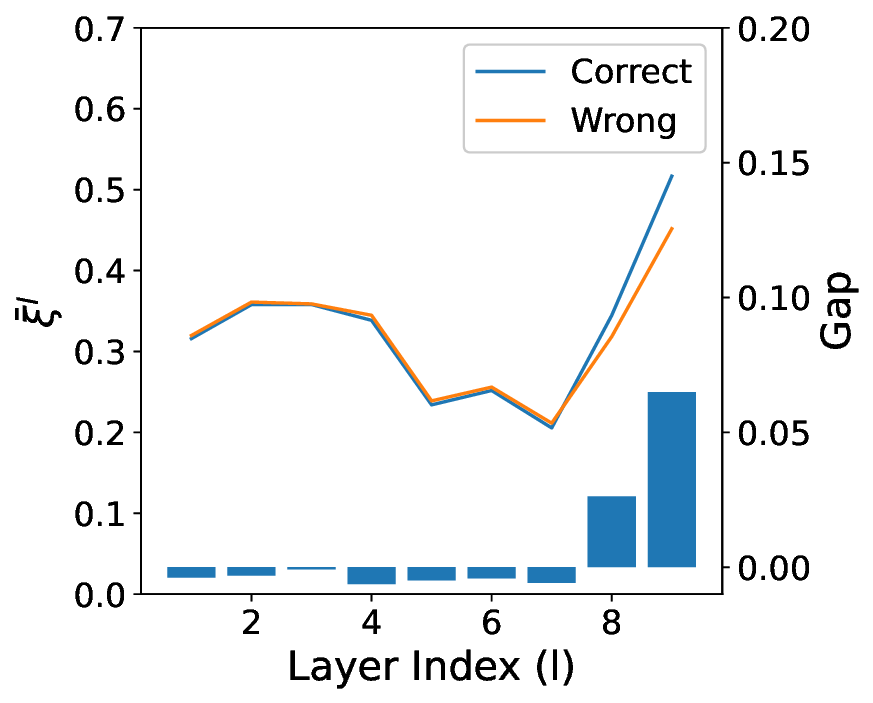}
    \caption{Regularised STF}
	\end{subfigure}
    \caption{Comparision of average STF $\bar \xi^l$ (over timestep) on Cifar10-DVS in distinguishing correct vs. wrong predictions. (a) illustrates the $\bar \xi^l$ using the baseline method, while (b) showcases the  $\bar \xi^l$ after applying the regularisation technique. In both cases, the left side shows the distribution of STF over different layers, and the right side displays the average STF values for correct and wrong predictions, alongside the gap value between them.}
    \label{fig:sft_vis}
\end{figure}

This section delves into the design of a regulariser based on the STF. Our prior analysis explained that TET prompts the network to minimise average loss without considering the sequential input order, leading to uncertain predictions at each timestep. To address this, we propose increasing the STF to decouple the influence of temporal information from network training. Nonetheless, this endeavor poses challenges, particularly in determining appropriate STF values for each layer without compromising accuracy. The right side of Figure \ref{fig:sft_vis} shows the STF distribution across different layers for both correct and wrong predictions. On the left side, we provide additional insights based on average values of STF with respect to the layer index. One noticeable observation is the progressive increase in STF as the layers grow deeper. Additionally, our experimental findings uncover a noteworthy discrepancy in STF values between correct and wrong predictions, particularly in the deeper layers. This observation indicates that wrong predictions tend to demonstrate lower STF values within these deeper layers. This intriguing phenomenon suggests that the SNN is actively involved in dampening spike occurrences when faced with difficult inputs.

To increase STF without sacrificing the accuracy, our goal is to eliminate the worst case -- the STF is relatively small -- during training. Thus, our regularisation only considers correct predictions during training, achieved by masking $\xi^l(t)$ as follows:

\begin{equation} \label{eq:phi-correct}
    \tilde \xi^l(t)  = 
    \left\{
    \begin{array}{ll}
        \xi^l(t) &  \textit{if correct prediction}\\
        0 &  \textit{else}.
    \end{array}
    \right. 
\end{equation}
Next, we assume that in an AOI-SNN, each timestep should contribute equally. To formalise this concept, we define $ \boldsymbol{\Xi}^l$ as a set of $ \tilde \xi^l(t)$ with all possible timesteps, expressed as:

\begin{equation} \label{eq:phi-transpose}
\boldsymbol{\Xi}^l =  [\tilde \xi^l(1),\tilde \xi^l(2), ..., \tilde \xi^l(T)]^\top.
\end{equation}

Then, STR is formulated as:
\begin{equation} \label{eq:KL}
R(\boldsymbol{\Xi}^l) = 
( \tilde  \xi^l_{min} - \tilde  \xi^l_{max} )^2,
\end{equation}
where $\tilde \xi_{min}^l$ and $\tilde \xi_{max}^l$ are the minimal and maximum values, respectively, in a mini-batch of $\boldsymbol{\Xi}^l$. Note that both values are non-zero so that the incorrect samples can be excluded during regularisation. We set $\tilde \xi_{max}$ to a relatively optimal value, because it is generally large while still ensuring correct predictions. 
To consider the STR across the total $L$ layers and adjust the loss function with the hyper-parameter $\alpha$, the final objective function becomes as:
\begin{equation} \label{eq:multiloss}
L_{TET} + \alpha \sum_{l}^L R(\boldsymbol{\Xi}^l),
\end{equation}

\section{Experiment}
In this section, we evaluate the effectiveness of our approach using uncertainty and synaptic operations as additional metrics alongside accuracy. Extensive experiments are conducted on frame-based and event-based datasets.

\subsection{Evaluation Metrics} \label{sec:synapticoperations}

\paragraph{Uncertainty Estimation} It is desired that the resulting SNN are always certain about its predictions at any timestep. 
Thus, we use the variance of predictions as a metric to evaluate the level of uncertainty. 
To quantify the uncertainty in prediction, we utilise the widely used ensemble method \cite{lakshminarayanan2017simple, fort2019deep} as it offers a straightforward approach for quantifying prediction uncertainty.
Specifically, we build an ensemble of SNN models, each trained from different weight initialisations. Such randomness in weight initialisations will lead the model to various solutions in the loss landscape, and therefore, the variance of predictions from the ensemble members reflect the uncertainty in predictions. 
Ensemble members are trained in parallel as they do not interact with each other. And during inference, the final prediction $\boldsymbol{\mu}(t)$ is the mean of predictions of all ensemble members: 
\begin{equation} \label{eq:var1}
\boldsymbol{\mu}(t) = \frac{1}{M}\sum^M_i f_i(\boldsymbol{X}(t)),
\end{equation}
where $M$ is the number of members in the ensemble, $t$ is the timestep. 
Then, the uncertainty or variance  of predictions at each timestep is calculated as: 
\begin{equation} \label{eq:var2}
\sigma^2(t) = \frac{1}{M}\sum^M_i \lVert f_i(\boldsymbol{X}(t)) - \boldsymbol{\mu}(t) \rVert_2,
\end{equation}
where larger $\sigma^2(t)$ implies higher uncertainty.

\paragraph{Synaptic Operations}  The energy efficiency of neuromorphic hardware can be characterised by the energy consumption of single synaptic operation \cite{Merolla:2014}. Thus, we follow \cite{rueckauer2017conversion, wu2022little} to measure the synaptic operations from simulation for energy consumption estimation, described as: 
\begin{align}
&\textit{Synaptic Operations}: \sum_{t}^T \sum_{l}^L f^l_{out}s^l,
\end{align}
where 
$f^l_{out}$ is the number of output connections and $s^l$ is the average number of spikes per neuron of the $l$-th layer.

\subsection{Experiment Setup}\label{app:experiment}
We evaluate SNN on ResNet-19 \cite{fang2021deep, deng2022temporal} for Cifar10/100 \cite{krizhevsky2009learning}, Sew-ResNet-34 \cite{fang2021deep} for ImageNet \cite{ILSVRC15}, VGGSNN \cite{deng2022temporal} for Cifar10-DVS  \cite{li2017cifar10dvs} and N-Caltech101 \cite{orchard2015converting}, 5-layer convoluational network \cite{fang2021incorporating} for DVS128 Gesture \cite{amir2017low}. 
Instead of rescaling the input, we adopt the approach from \cite{wu2023optimising}, in which we incorporate a downscaling layer prior to the network. Specifically, for Cifar10-DVS and N-Caltech101, we add a convolutional layer comprising 64 filters, a kernel size of 8, and strides of 4. The number of filters is increased to 128 for DVS128 Gesture. This modification allows the events to be directly fed into the SNN while preserving the event-driven features.

We employ Stochastic Gradient Descent (SGD) with an initial learning rate of 0.1 and weight decay of 5e-4 for all datasets. The training epochs are set to 300, 120, and 100 for Cifar10/100, ImageNet, and event-based datasets, respectively. The learning rate is decayed to zero at the end of training using the cosine decay schedule \cite{loshchilov2016sgdr}.
For data augmentation, we use autoaugmentation \cite{cubuk2019autoaugment} and cutout \cite{devries2017improved} for Cifar10/100 and pixel shifting for event-based inputs, i.e., both width and height are randomly shifted by the range [-20\%,20\%]. Dropout is applied after fully-connected layer for DVS128 Gesture to improve the training and the dropout rate is 0.2. In the results, we use `TET' to present the baseline \cite{deng2022temporal} and `STR($\cdot$)' to denote our method with the setting of $\alpha$ in the bracket. We follow 'TET' to adopt the surrogate method described in \cite{esser2015backpropagation}. Since the direct training is extraordinary expense for the training time, we train 3 models for ImageNet and 5 models for the other datasets with different seeds so that $M$ is 3 or 5 for Equation \ref{eq:var1} and \ref{eq:var2}.

\subsection{Uncertainty Results}



\begin{figure*}[t!]
    \centering
    \begin{subfigure}[t]{0.33\textwidth}
    \includegraphics[width=\linewidth]{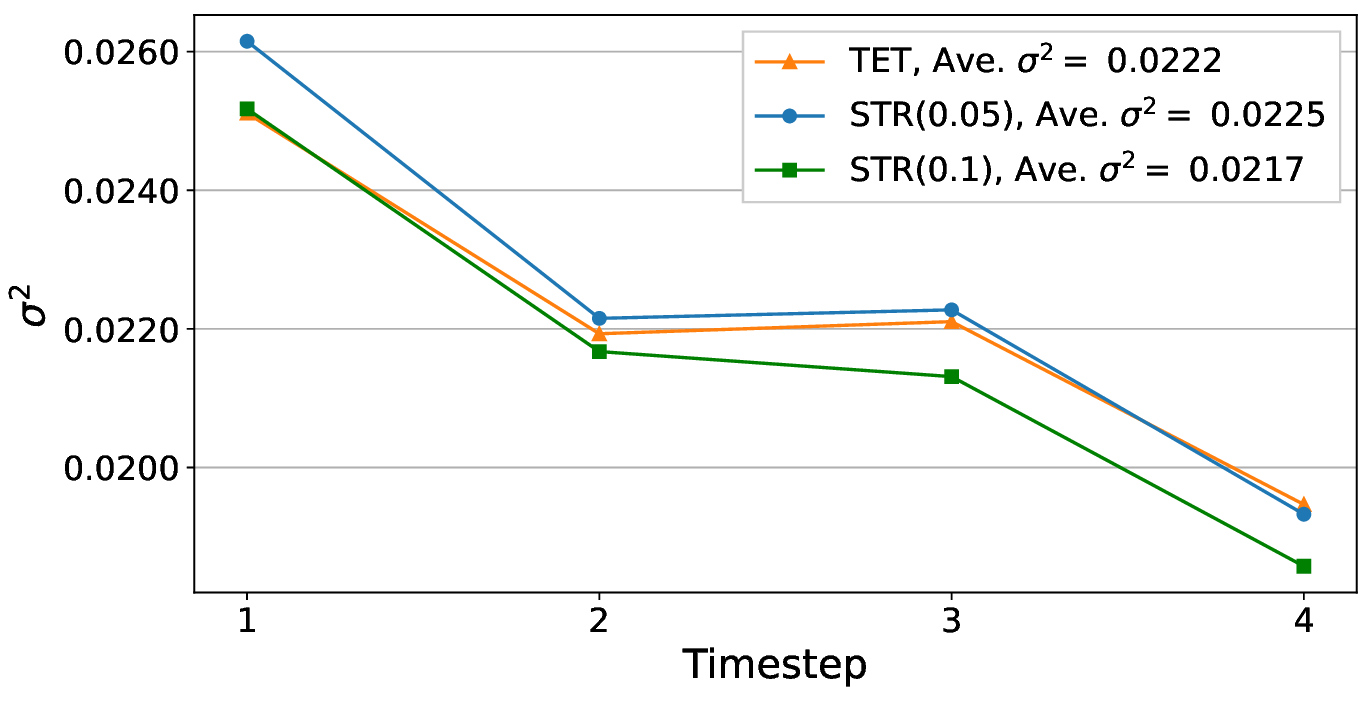}
    \caption{Cifar10}
    
    \end{subfigure}
    \begin{subfigure}[t]{0.33\textwidth}
    \includegraphics[width=\linewidth]{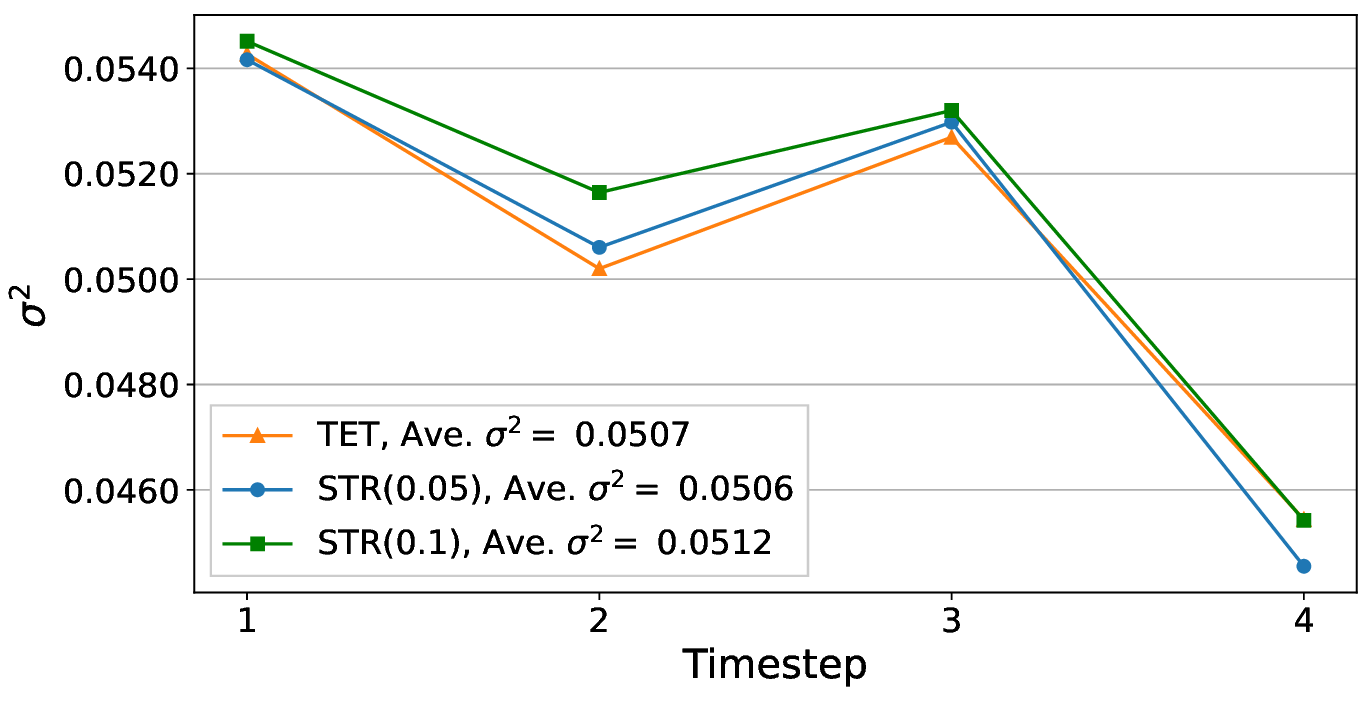}
    \caption{Cifar100}
    
	\end{subfigure}
    \begin{subfigure}[t]{0.33\textwidth}
    \includegraphics[width=\linewidth]{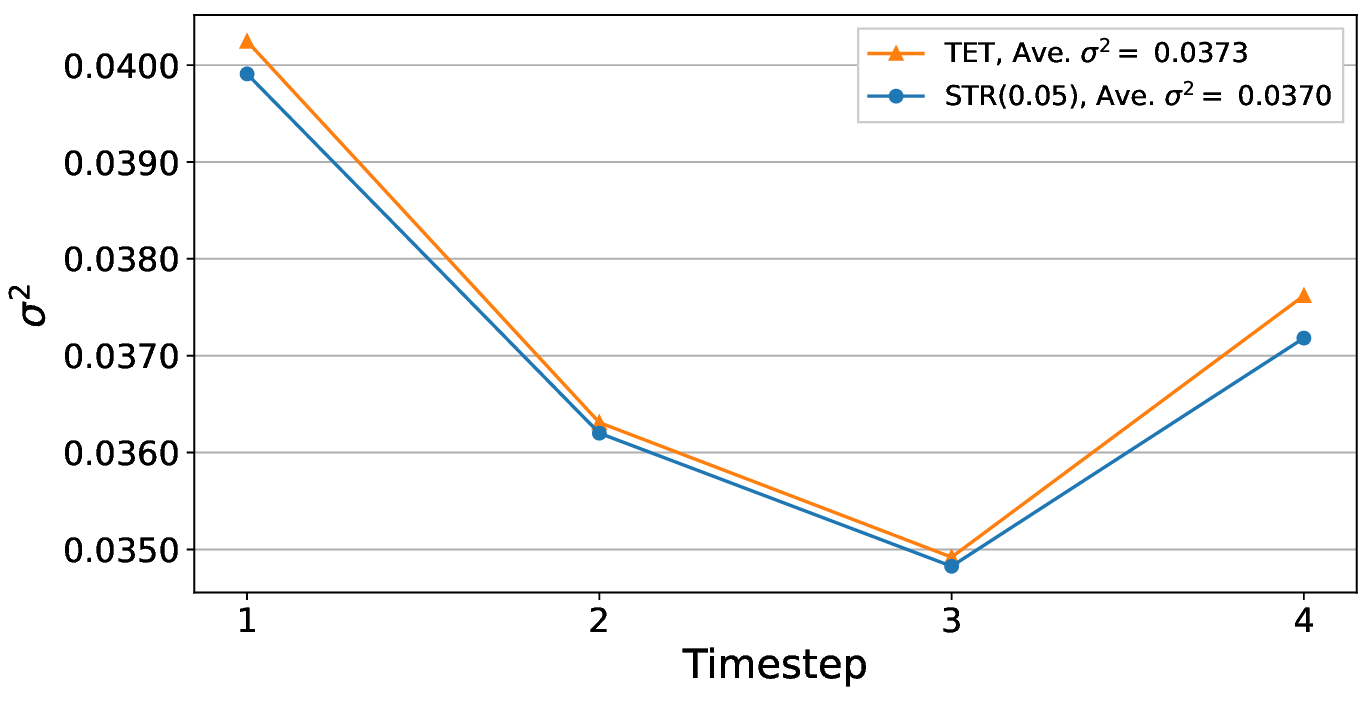}
    \caption{ImageNet}
    
	\end{subfigure}
    \begin{subfigure}[t]{0.33\textwidth}
    \includegraphics[width=\linewidth]{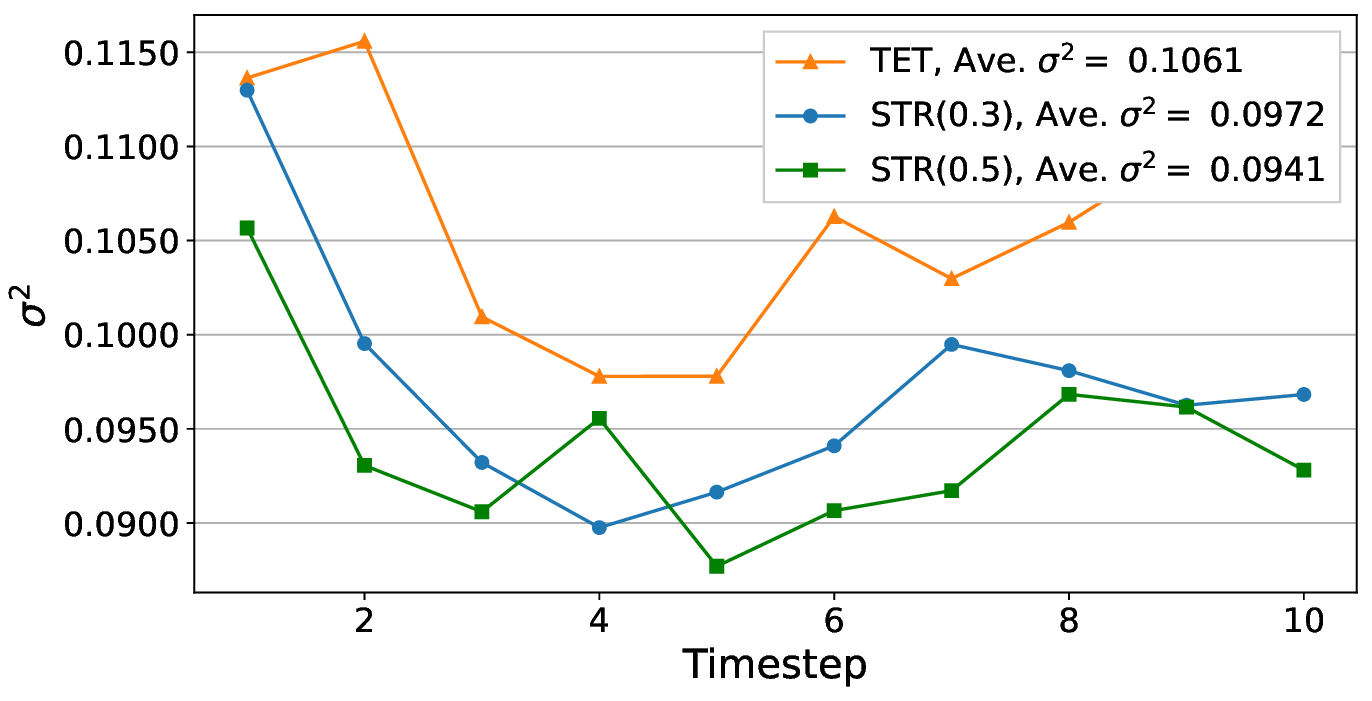}
    \caption{Cifar10-DVS}
    
	\end{subfigure}
     \begin{subfigure}[t]{0.33\textwidth}
    \includegraphics[width=\linewidth]{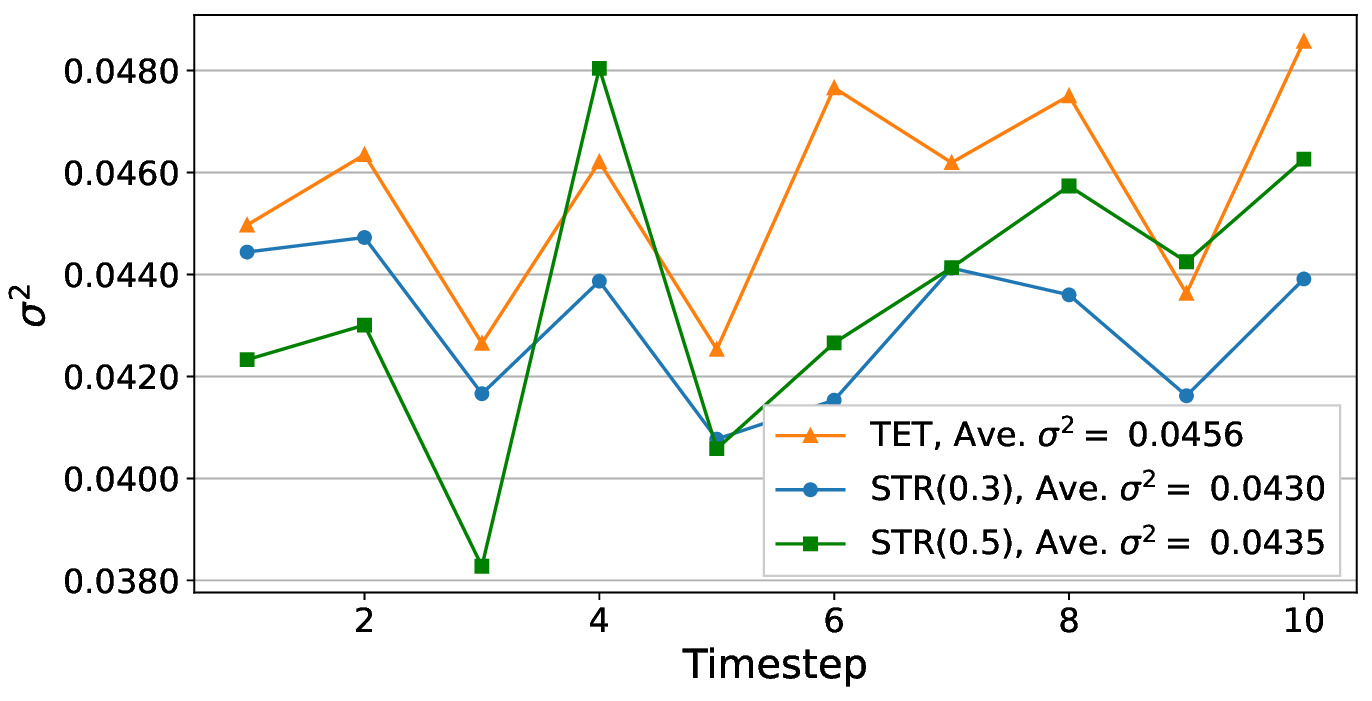}
    \caption{N-Caltech101}
    \label{fig:ncaltehc101-var}
	\end{subfigure}
    \begin{subfigure}[t]{0.33\textwidth}
    \includegraphics[width=\linewidth]{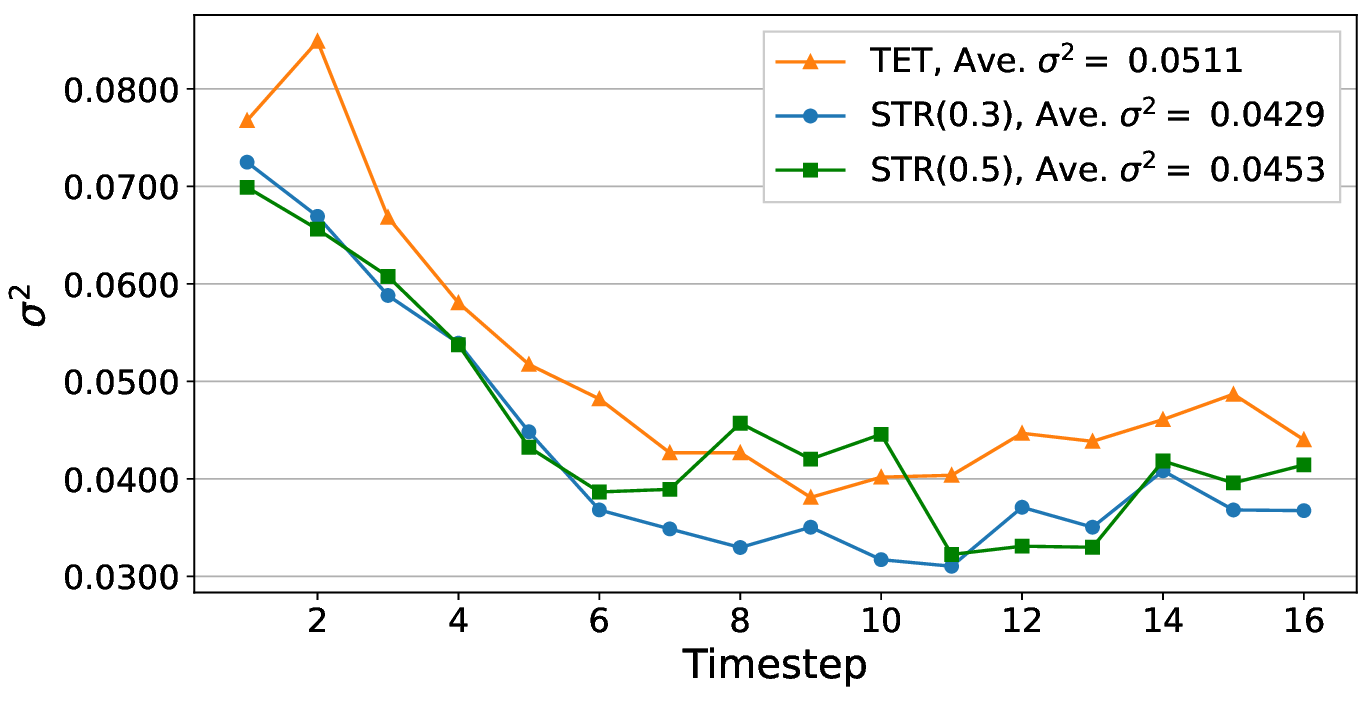}
    \caption{DVS128 Gesture}
    
	\end{subfigure}
	\caption{Comparison of uncertainty with respect to timestep on six datasets. The assessment of uncertainty is conducted on various models with distinct settings of $\alpha$, such as \{0.05, 0.1\} for Cifar10/100 and \{0.3, 0.5\} for event-based inputs. Due to the expensive training for ImageNet, we solely evaluate the setting of \{0.05\}.
    }
	\label{fig:var_all}
\end{figure*}


Figure \ref{fig:var_all} depicts the estimated uncertainty of predictions on different datasets, which reveals interesting patterns of uncertainty trends as time evolve. Specifically, it can be observed that the predictions at initial timestep tend to exhibit large uncertainty on most datasets, and then gradually reduces. This shows predictions becomes more reliable as the timestep increases. But we also notice that such trend is not obvious on the N-Caltech101 dataset, as shown in Figure \ref{fig:ncaltehc101-var}.
This is because the N-Caltech101 dataset has more events, providing more useful information for classification. For example, N-Caltech101 has an average of 5230 spikes per second for the input, while Cifar10-DVS only has 85.38 spikes per second. By incorporating STR, we observe a significant decrease in uncertainty in predictions, which implies improved stability and reliability in the predictions throughout the temporal sequence. Table \ref{tab:var_avg} summarises the average uncertainty and the accuracy achieved at the last timestep over all datasets. The results indicate that training with STR consistently decreases uncertainty while maintaining accuracy or even achieving higher accuracy for event-based datasets.

\begin{table}
    \caption{Comparison of average uncertainty and accuracy at the last timestep on both frame-based and event-based datasets. The optimal $\alpha$ for each model is selected based on Figure \ref{fig:var_all}, prioritising relatively small variance while preserving accuracy.}
    \label{tab:var_avg}
    \centering
    \small
    \begin{tabular*}{\linewidth}{@{\extracolsep{\fill}} c c c c c}
    \Xhline{4\arrayrulewidth}
    Dataset  & Method &T & Avg. $\sigma^2 $ &Avg. Acc. (\%) \\
    \Xhline{4\arrayrulewidth}
     \multirow{2}{6em}{Cifar10}
     & TET  &4 &0.0222 & 95.40 $\pm$ 0.05 \\
     & \textbf{STR(0.1)} &4  & \textbf{0.0217} & \textbf{95.42 $\pm$ 0.04} \\
    \hline
     \multirow{2}{6em}{Cifar100}
     & TET &4 & 0.0507 & 78.31 $\pm$ 0.14 \\
     & \textbf{STR(0.05)} &4 & \textbf{0.0506} & \textbf{78.37 $\pm$ 0.22} \\
    \hline
     \multirow{2}{6em}{ImageNet}
     & TET &4 & 0.0373 & 67.46 $\pm$ 0.02 \\
     & \textbf{STR(0.05)} &4 & \textbf{0.0370} & \textbf{67.54 $\pm$ 0.03} \\
    \Xhline{4\arrayrulewidth}
     \multirow{2}{6em}{Cifar10-DVS}
     & TET &10 & 0.1061 & 82.38 $\pm$ 0.59 \\
     & \textbf{STR(0.5)} &10 & \textbf{0.0941} & \textbf{82.64 $\pm$ 0.44} \\
    \hline
     \multirow{2}{6em}{N-Caltech101}
     & TET &10 & 0.0456 & 84.84 $\pm$ 0.41 \\
     & \textbf{STR(0.5)}  &10 & \textbf{0.0435} & \textbf{85.91 $\pm$ 0.54} \\
    \hline
    \multirow{2}{6em}{DVS128 Gesture}
     & TET &16 & 0.0511 & 97.80 $\pm$ 0.37 \\
     & \textbf{STR(0.5)} &16 & \textbf{0.0453} & \textbf{98.26 $\pm$ 0.30} \\
   
    \Xhline{4\arrayrulewidth}           
    \end{tabular*}

\end{table}



\subsection{Cutoff Results}

\begin{figure*}[h!]
    \centering
    \begin{subfigure}[t]{0.33\textwidth}
    \includegraphics[width=\linewidth]{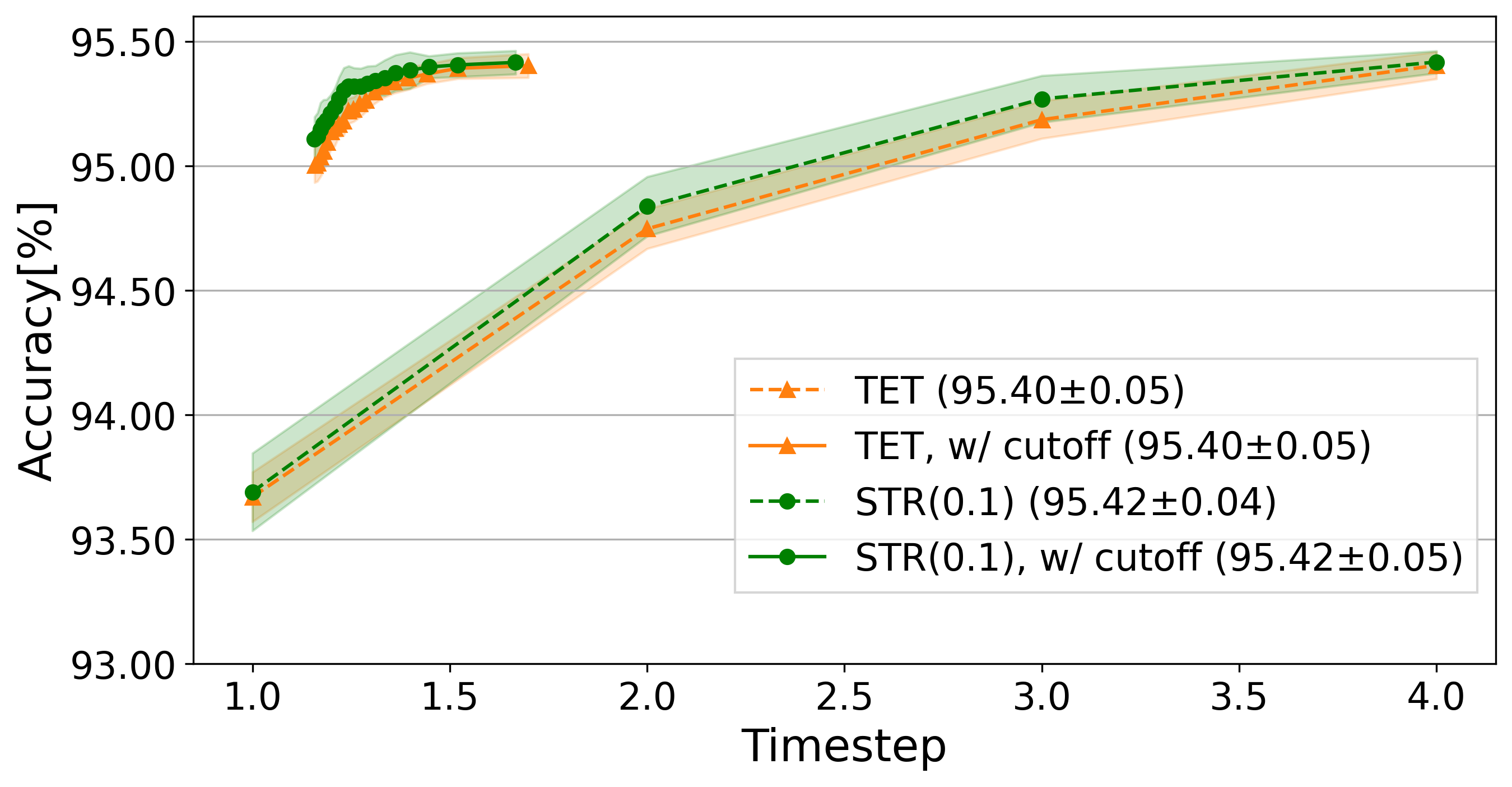}
    \caption{Cifar10}
    \label{fig:cutoff-a}
    \end{subfigure}
    \begin{subfigure}[t]{0.33\textwidth}
    \includegraphics[width=\linewidth]{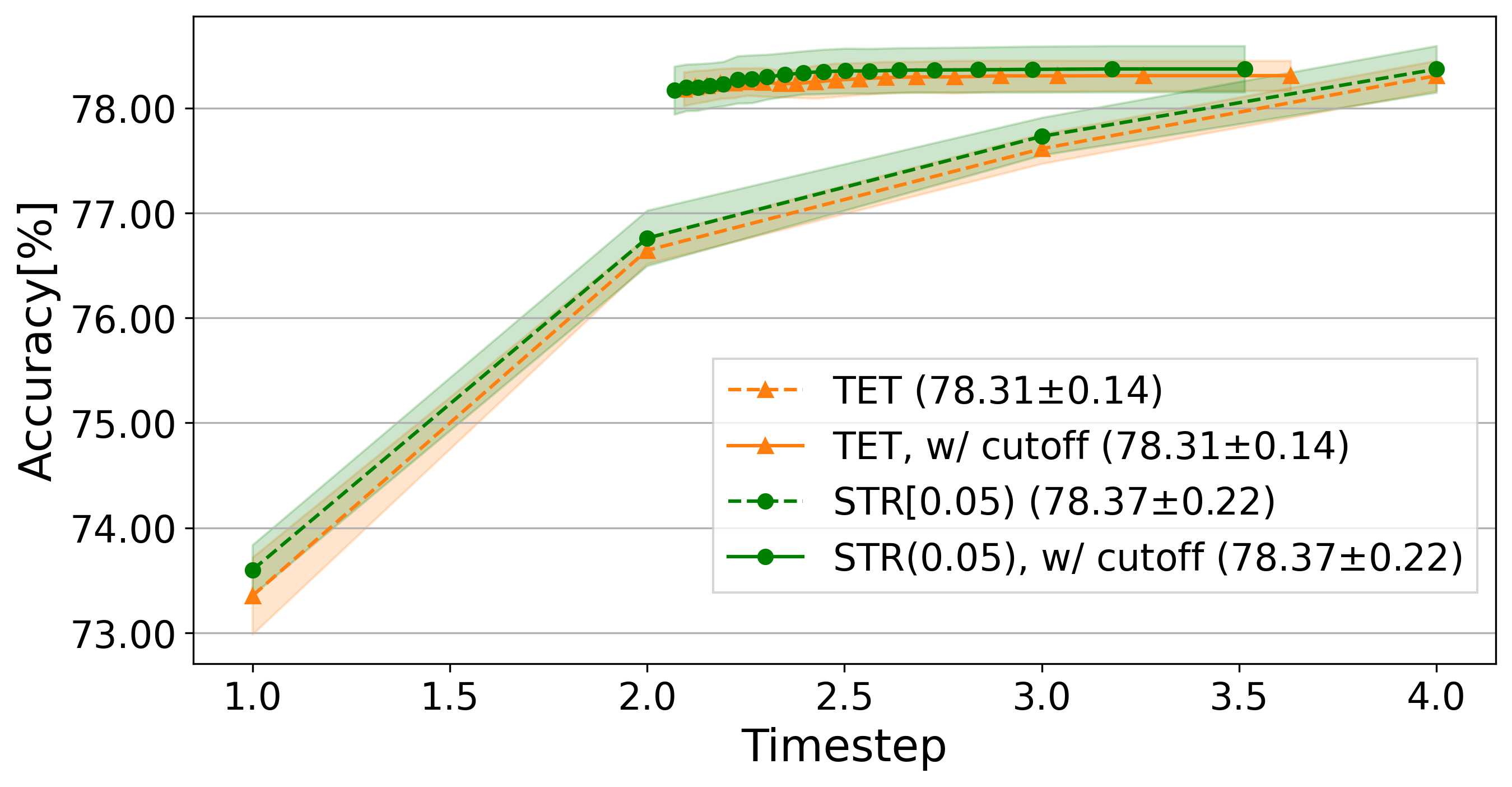}
    \caption{Cifar100}
    \label{fig:cutoff-b}

	\end{subfigure}
     \begin{subfigure}[t]{0.33\textwidth}
    \includegraphics[width=\linewidth]{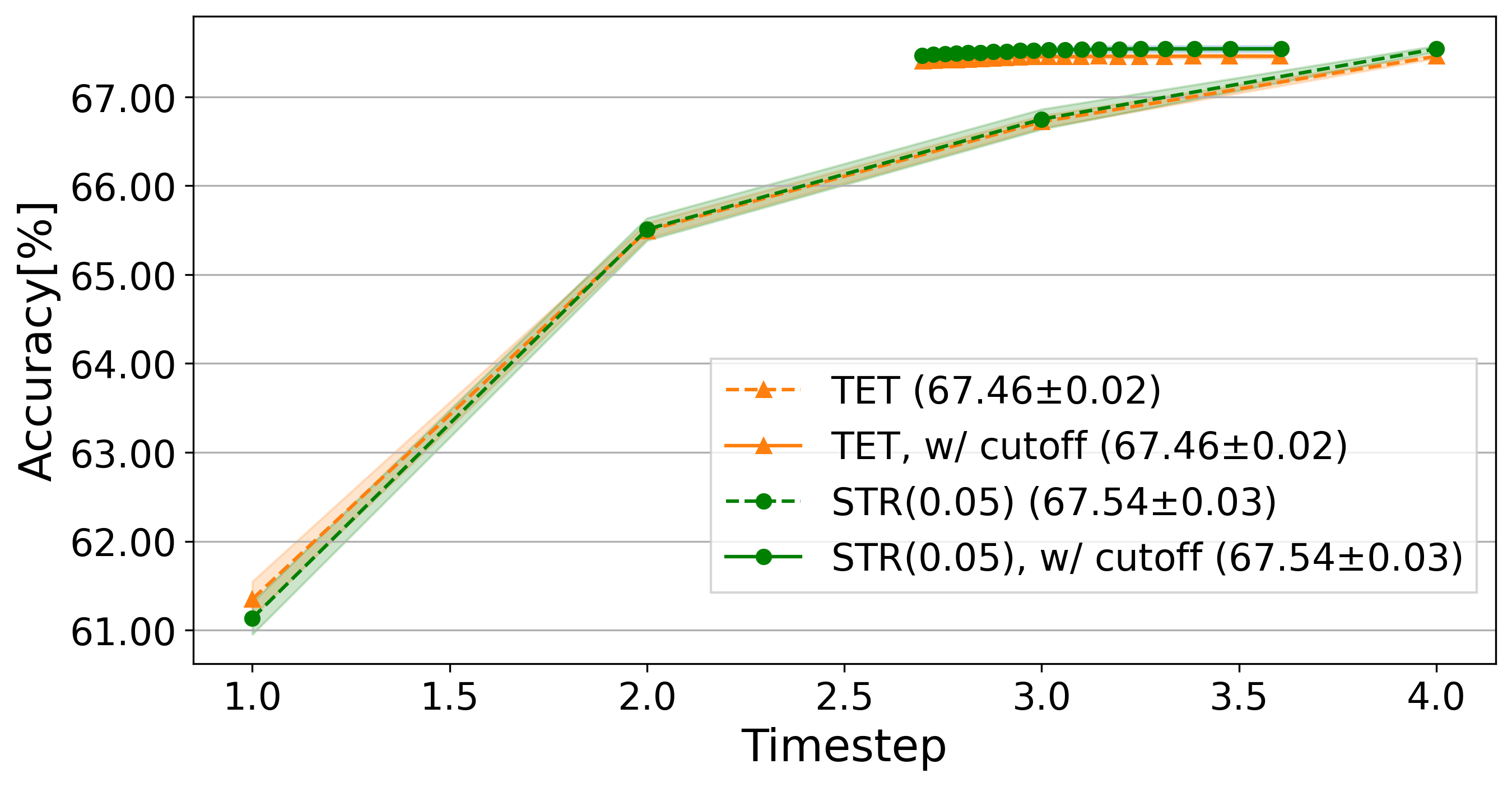}
    \caption{ ImageNet}
    \label{fig:cutoff-c}

	\end{subfigure}
    \begin{subfigure}[t]{0.33\textwidth}
    \includegraphics[width=\linewidth]{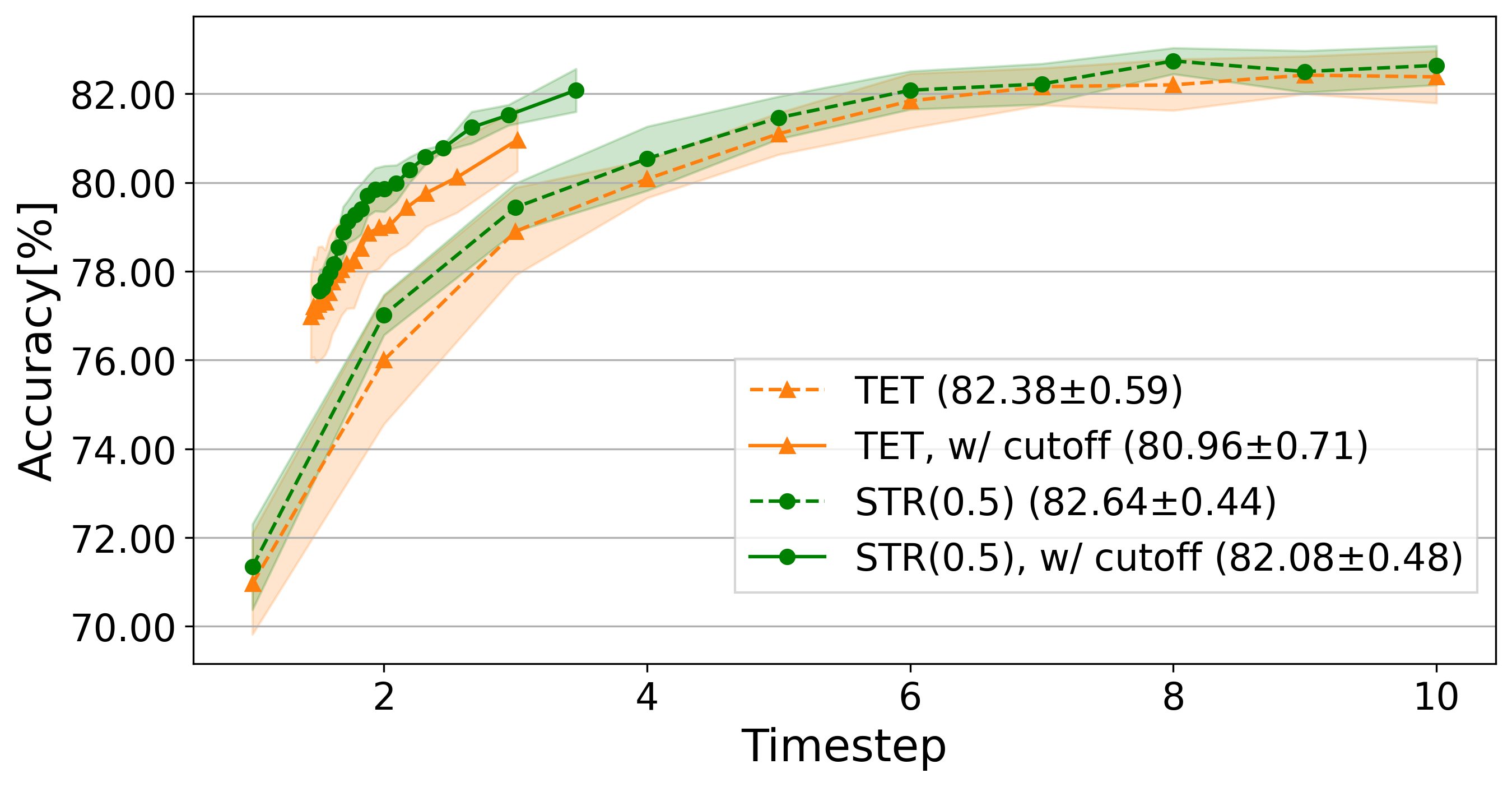}
    \caption{Cifar10-DVS}
    \label{fig:cutoff-d}

	\end{subfigure}
     \begin{subfigure}[t]{0.33\textwidth}
    \includegraphics[width=\linewidth]{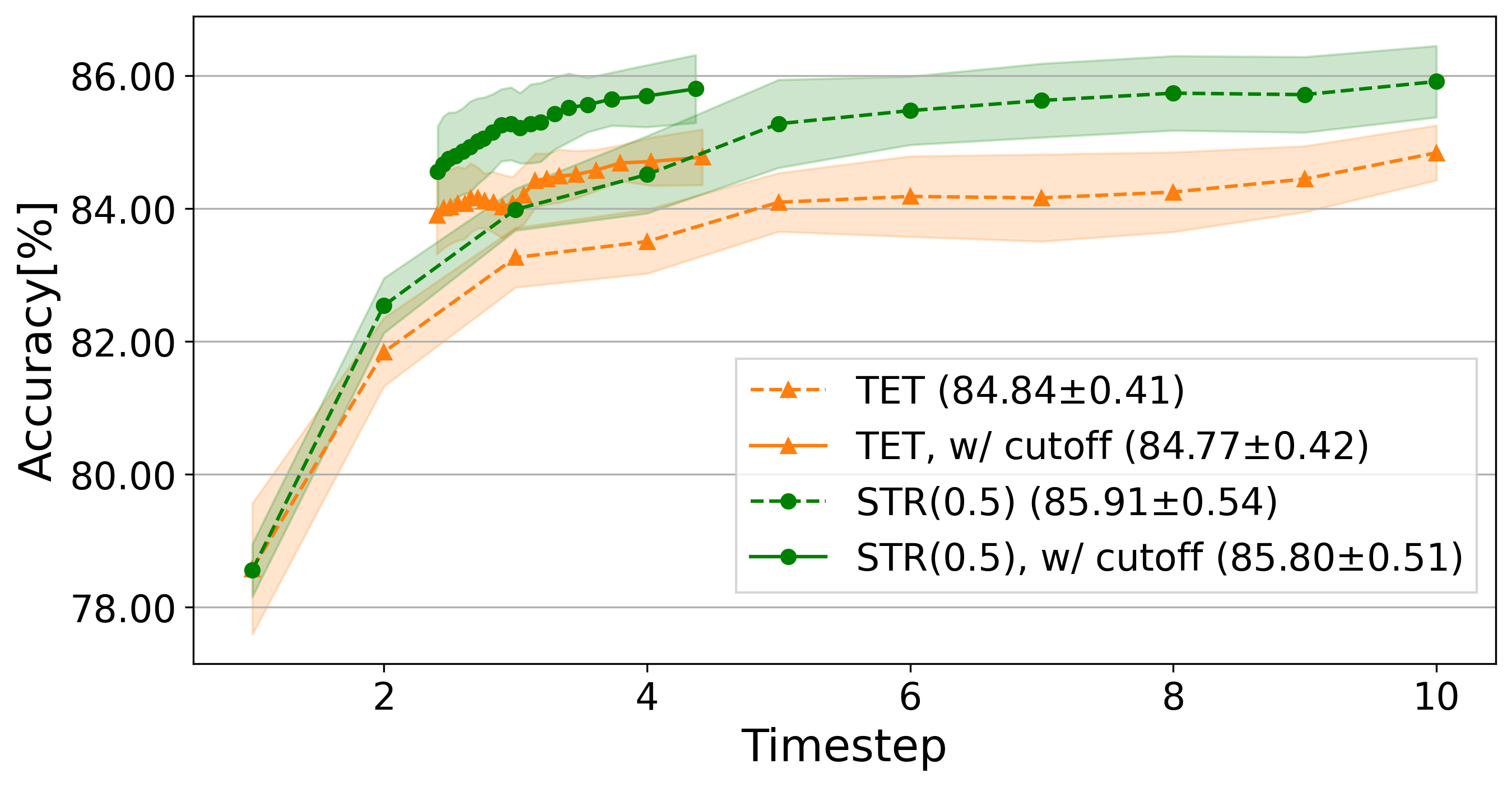}
    \caption{N-Caltech101}
    \label{fig:cutoff-e}

	\end{subfigure}
    \begin{subfigure}[t]{0.33\textwidth}
    \includegraphics[width=\linewidth]{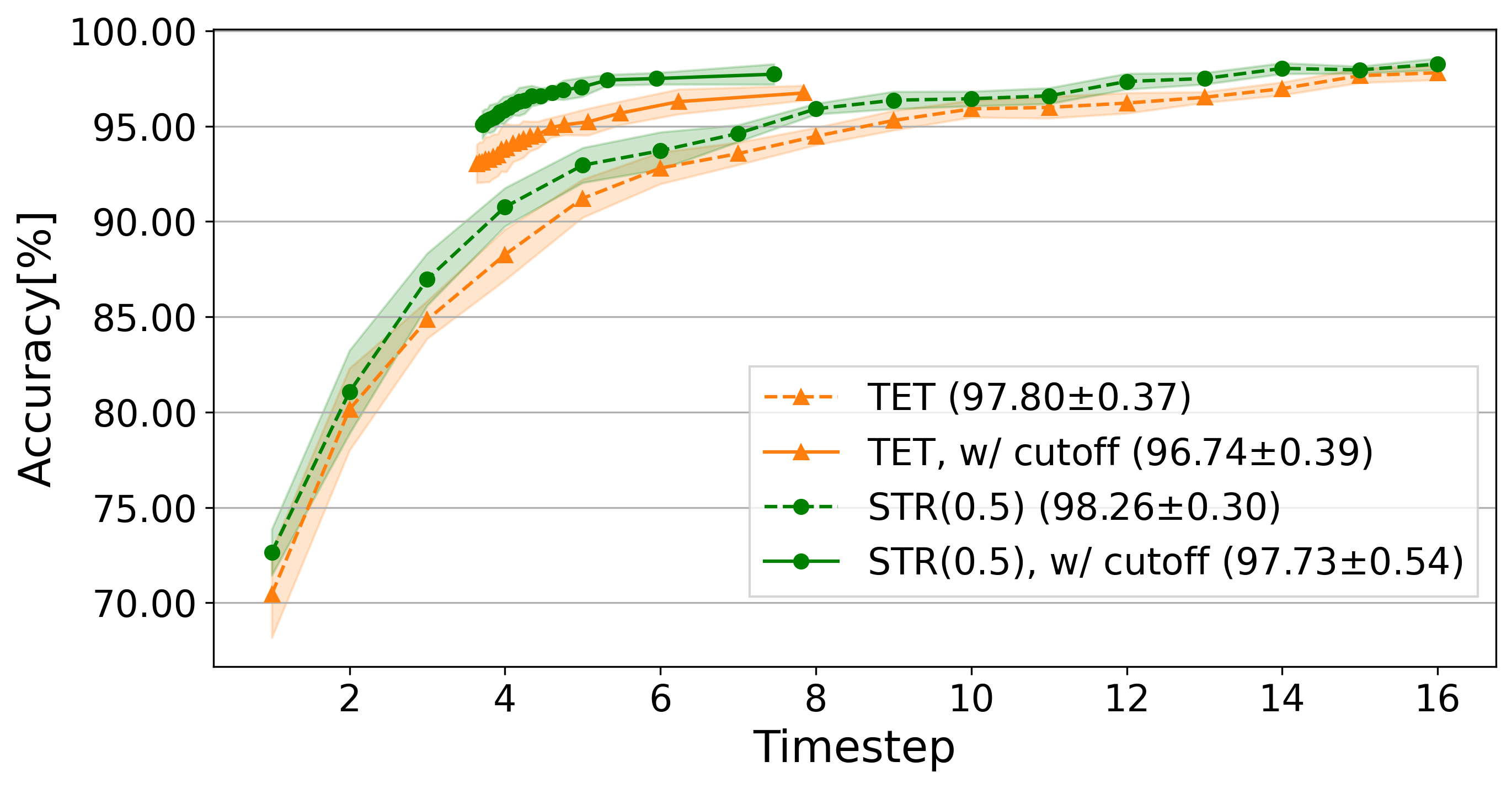}
    \caption{DVS128 Gesture}
    \label{fig:cutoff-f}

	\end{subfigure}
 %
 %
 %
	\caption{Comparison of accuracy with respect to timestep on six datasets. Each STR-based model employs the same $\alpha$ setting from Table \ref{tab:var_avg}. The results present the accuracy performance using fixed timesteps and cutoff.}
	\label{fig:cutoff}
\end{figure*}

As previously highlighted, using STR can be effective in reducing the variance across the timesteps, especially on event-based dataset in which the maximum timestep is relatively large for training. Further insights into the comparison between TET and STR are illustrated in Figure \ref{fig:cutoff}, considering two different inference types -- one with fixed timestep and the other one with cutoff mechanism. The label `w/ cutoff' signifies results with cutoff.  While the curve without cutoff has often been utilised to find a balance between timesteps and accuracy \cite{Han:2020, wu2022little}, it has a drawback of fixing the timestep during inference, leading to a notable decline in accuracy when the timestep is small. As TET trains SNN to predict at each timestep, we directly apply softmax-based cutoff on the resulted SNN models for anytime inference. Precisely, the SNN is cutoff when the maximum softmax score at the output surpasses the predetermined threshold.

Figure \ref{fig:cutoff} presents accuracy with respect to a range of cutoff threshold varies from [0.99 to 1.0] for the DVS128 Gesture, and [0.8 to 1.0] for the other datasets. In both instances, the threshold range is divided into 20 discrete values, each having an equal interval between them. It shows that with cutoff all models have significant decrease on the latency while maintaining the accuracy. Compared to frame-based input (e.g., Figure \ref{fig:cutoff-a} to \ref{fig:cutoff-c}), the enhancement from STR in event-based input (e.g., Figure \ref{fig:cutoff-d} to \ref{fig:cutoff-f}) is more substantial. This is attributed to the sparser nature and greater uncertainty in predictions associated with event input. In contrast, frame-based input data furnishes more information at each timestep, aiding in the prediction.

\begin{table}[!ht]
    \caption{Comparison with the exiting works on frame-based datasets in regard to both accuracy and latency.}
    \label{tab:cutoff-1}
    \begin{subtable}[h]{\linewidth}
    \centering
    \small
    \begin{tabular*}{\linewidth}{@{\extracolsep{\fill}}  c  c c c}
    \Xhline{4\arrayrulewidth}
    Methods  & Architecture  &Avg. Acc. (\%) & Avg. $T$ \\
    \Xhline{4\arrayrulewidth}
     \cite{li2021differentiable} & ResNet-18  & 93.13$\pm$0.07 & 2\\
     \cite{zheng2021going} & ResNet-19  & 92.92 & 4\\
     \cite{yao2022glif} & ResNet-19  & 94.44$\pm$0.10 & 2\\
     \cite{duan2022temporal} & ResNet-19  & 95.45 & 2\\

    \Xhline{4\arrayrulewidth}
      TET@0.9 &\multirow{2}{4.3em}{ResNet-19}   & 95.23  $\pm$ 0.05 &  1.26  \\ 
      \textbf{STR@0.9} &    & \textbf{95.32$\pm$ 0.07} &  \textbf{1.26} \\ 
    \hline

      TET@1.0  &\multirow{2}{4.3em}{ResNet-19}   & 95.40$\pm$ 0.05 &  1.70  \\ 
      \textbf{STR@1.0}  &   & \textbf{95.42 $\pm$ 0.05} &  \textbf{1.67} \\ 
    \Xhline{4\arrayrulewidth}           
    \end{tabular*}
    \caption{Cifar10}
    \end{subtable}
    \begin{subtable}[h]{\linewidth}
    \centering
    \small
    \begin{tabular*}{\linewidth}{@{\extracolsep{\fill}}  c  c c c}
    \Xhline{4\arrayrulewidth}
    Methods  & Architecture  &Avg. Acc. (\%) & Avg. $T$ \\
    \Xhline{4\arrayrulewidth}
     \cite{li2021differentiable} & ResNet-18  & 71.68$\pm$0.12 & 2\\
     \cite{deng2022temporal} & ResNet-19  & 72.87$\pm$0.10 & 2\\
     \cite{yao2022glif} & ResNet-19  & 75.48$\pm$0.08 & 2\\
     \cite{duan2022temporal} & ResNet-19  & 78.07 & 2\\
    \Xhline{4\arrayrulewidth}
      TET@0.9  &\multirow{2}{4.3em}{ResNet-19}  & 78.25  $\pm$ 0.17 &  2.42  \\ 
       \textbf{STR@0.9}  &   & \textbf{78.34$\pm$ 0.21} &  \textbf{2.40} \\ 
    \hline
      TET@1.0  &\multirow{2}{4.3em}{ResNet-19}  & 78.31$\pm$ 0.14 &  3.63  \\ 
      \textbf{STR@1.0}  &    & \textbf{78.37 $\pm$ 0.22} &  \textbf{3.51} \\ 
    \Xhline{4\arrayrulewidth}           
    \end{tabular*}
    \caption{Cifar100}
    \end{subtable}
    \hfill
    \begin{subtable}[h]{\linewidth}
    \small
    \centering
    \begin{tabular*}{\linewidth}{@{\extracolsep{\fill}}  c  c c c}

    \Xhline{4\arrayrulewidth}
    Methods  & Architecture  &Avg. Acc. (\%) & Avg. $T$ \\
    \Xhline{4\arrayrulewidth}
     \cite{bu2022optimal} & ResNet-34  & 59.35 & 16\\
     \cite{meng2022training} & ResNet-34  & 67.05 & 6\\
     \cite{fang2021deep} & \scriptsize{SewResNet-34}  & 67.04 & 4\\
     \cite{duan2022temporal} & \scriptsize{SewResNet-34}  & 68.28 & 4\\
    \Xhline{4\arrayrulewidth}
      TET@0.8  &\multirow{2}{4.8em}{\scriptsize{SewResNet-34}}   & 67.40  $\pm$ 0.02 &  2.70  \\ 
       \textbf{STR@0.8}  &   & \textbf{67.46$\pm$ 0.04} &  \textbf{2.70} \\ 
    \hline
      TET@1.0  &\multirow{2}{4.8em}{\scriptsize{SewResNet-34}}   & 67.46$\pm$ 0.02 &  3.60  \\ 
      \textbf{STR@1.0}  &     & \textbf{67.54 $\pm$ 0.03} &  \textbf{3.60} \\ 
    \Xhline{4\arrayrulewidth}           
    \end{tabular*}
    \caption{ImageNet}
    \end{subtable}

\end{table}

\begin{table}[!ht]
    \caption{Comparison with the exiting works on event-based datasets in regard to both accuracy and latency. \textbf{Note:} * to indicate the network with downscaling layer. }
    \label{tab:cutoff-2}
    \begin{subtable}[h]{\linewidth}
    \centering
    \small
    \begin{tabular*}{\linewidth}{@{\extracolsep{\fill}}  c  c c c}
    \Xhline{4\arrayrulewidth}
    Methods  & Architecture  &Avg. Acc. (\%) & Avg. $T$ \\
    \Xhline{4\arrayrulewidth}
     \cite{fang2021deep} &  Wide-7B-Net  & 74.40 & 16\\
     \cite{meng2022training} &  ResNet-19  & 67.80 & 10\\
      \cite{li2021differentiable} & ResNet-18  & 75.4$\pm$0.05 & 10\\
      \cite{li2022neuromorphic} & VGG-11  & 81.70 & 10\\
    \Xhline{4\arrayrulewidth}
      TET@0.9  &\multirow{2}{4em}{VGGSNN*}   & 78.24 $\pm$ 0.99 &  1.71 \\
      \textbf{STR@0.9}  &   & \textbf{79.70 $\pm$ 0.20} &  \textbf{1.88} \\ 
    \hline
      TET@1.0 &\multirow{2}{4em}{VGGSNN*}   & 80.96$\pm$ 0.50 &  3.01  \\ 
      \textbf{STR@1.0}  &   & \textbf{82.08 $\pm$ 0.23} &  \textbf{3.46} \\ 
    \Xhline{4\arrayrulewidth}           
    \end{tabular*}
    \caption{Cifar10-DVS}
    \end{subtable}
    \begin{subtable}[h]{\linewidth}
    \centering
    \small
    \begin{tabular*}{\linewidth}{@{\extracolsep{\fill}}  c  c c c}
    \Xhline{4\arrayrulewidth}
    Methods  & Architecture  &Avg. Acc. (\%) & Avg. $T$ \\
    \Xhline{4\arrayrulewidth}
     \cite{li2021graph} & Graph   & 76.10 & - \\
     \scriptsize{\cite{messikommer2020event}} & VGG-13 & 74.50 & - \\
     \cite{li2022neuromorphic} & VGG-11 & 83.70 & 10 \\
    \Xhline{4\arrayrulewidth}
      TET@0.9  &\multirow{2}{4em}{VGGSNN*}   & 84.07 $\pm$ 0.16 &  2.98 \\
      \textbf{STR@0.9}  & & \textbf{85.27 $\pm$ 0.30} &  \textbf{2.97} \\ 
    \hline
      TET@1.0 &\multirow{2}{4em}{VGGSNN*}   & 84.77$\pm$ 0.17 &  4.41  \\ 
      \textbf{STR@1.0}  &  & \textbf{85.80 $\pm$ 0.26} &  \textbf{4.37} \\ 
    \Xhline{4\arrayrulewidth}           
    \end{tabular*}
    \caption{N-Caltech101}
    \end{subtable}
    \hfill
    \begin{subtable}[h]{\linewidth}
    \small
    \centering

    \begin{tabular*}{\linewidth}{@{\extracolsep{\fill}}  c  c c c}

    \Xhline{4\arrayrulewidth}
    Methods  & Architecture  &Avg. Acc. (\%) & Avg. $T$ \\
    \Xhline{4\arrayrulewidth}
     \cite{yao2021temporal} & 3-layer & 98.61 & 60 \\
     \cite{zheng2021going} & ResNet-17  & 96.87 & 40\\
     \cite{fang2021incorporating} & 5-layer  & 97.46 & 20\\
     \cite{fang2021deep} & 7B-Net  & 97.92 & 16\\
    \Xhline{4\arrayrulewidth}
      TET@0.99   &\multirow{2}{4em}{5-layer*}  & 94.16 $\pm$ 0.92 &  4.18  \\ 
      \textbf{STR@0.99} &  & \textbf{96.29 $\pm$ 0.73} &  \textbf{4.18} \\ 
    \hline
      TET@1.00  &\multirow{2}{4em}{5-layer*}   & 96.74$\pm$ 0.37 &  7.85 \\ 
     \textbf{STR@1.00} &    & \textbf{97.73 $\pm$ 0.54} &  \textbf{7.46} \\ 
    \Xhline{4\arrayrulewidth}           
    \end{tabular*}
    \caption{DVS128 Gesture}
    \end{subtable}

\end{table}



For a comprehensive comparison with state-of-the-art (SOTA) methods, we summarise accuracy data based on different cutoff thresholds using the format \textit{method@cutoff threshold} in Table \ref{tab:cutoff-1} and \ref{tab:cutoff-2}.
Note that our objective is not to achieve higher accuracy than SOTA models, but rather to achieve comparable accuracy while reducing the average timesteps required for inference. By examining the results in Table \ref{tab:cutoff-1} and \ref{tab:cutoff-2}, it is evident that our STR approach consistently achieves competitive latency with the SOTA and baseline. 
Notably, when employing these techniques, SNN showcases a remarkable acceleration in inference times. With STR and cutoff, SNN achieves $2.14$ to $2.89$ times faster in inference compared to SNN presented in Table \ref{tab:var_avg}, which uses a fixed timestep.
This enhanced efficiency is achieved with a near-zero accuracy drop of $0.50\%$ to $0.64\%$ over the event-based datasets. 

\begin{wrapfigure}{r}{0.46\textwidth}
    \begin{center}
  \includegraphics[width=\linewidth]{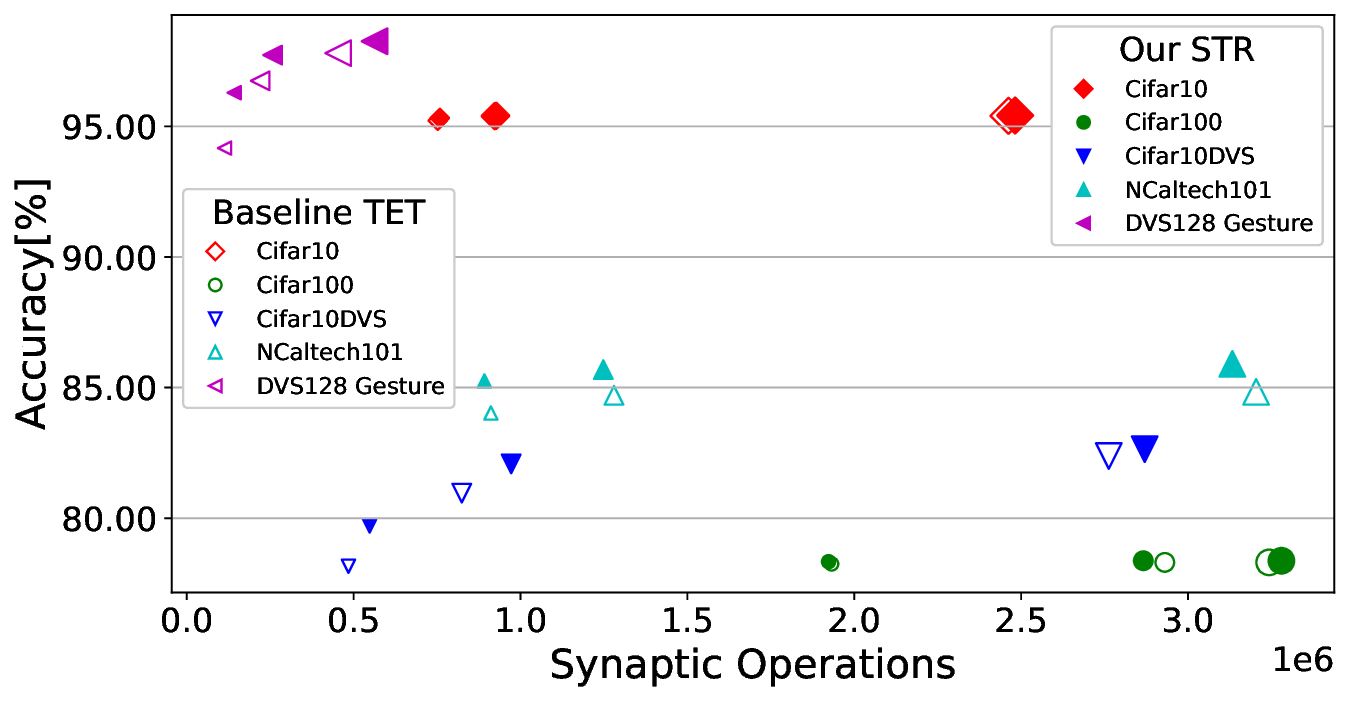}
    \caption{Comparison of accuracy with respect to synaptic operations.  The cutoff threshold is set to  \{0.99, 1.0, $\infty$\} for DVS128 Gesture and \{0.9, 1.0, $\infty$\} for the others, i.e., smaller marker indicates lower cutoff threshold and $\infty$ means no cutoff.}
    \label{fig:mops_comparison}
    \end{center}
\end{wrapfigure}

Figure \ref{fig:mops_comparison}, which illustrates accuracy relative to synaptic operations, providing clear evidence that cutoff significantly diminishes the number of synaptic operations required. It should be noted that the count of synaptic operations for Cifar10/100 excludes the first layers, which is a traditional ANN layer and function as a spike encoder. STR may introduce additional spikes to enhance accuracy. For instance, in the case of DVS128 Gesture, the number of synaptic operations is higher for STR compared to TET. However, when comparing reduced synaptic operations from cutoff, the increment is marginal. 

Following the presentation of our experimental results, it is pertinent to contextualise these findings within the broader scope of related work, specifically in comparison to SEENN \cite{li2023seenn}. Our baseline `TET', akin to SEENN-I, is a direct implementation of the methods outlined in \cite{deng2022temporal} and with utilising the confidence score for the cutoff. Thus, this prompts us to focus the comparison over our baseline, which shares identical training settings. Comparing with SEENN-II, which requires an additional network to trigger the cutoff, our STR focus on optimising the SNN itself, maintaining its original structure and enhancing performance in cutoff. This approach provides an alternative path in SNN advancements, distinct from and not in conflict with external strategy like SEENN-II.

\section{Conclusion}
In light of the approach presented in our work, we have demonstrated the effectiveness of STR in enhancing the reliability of SNN for anytime inference scenarios. Combining with the cutoff mechanism, our approach further enhances the performance metrics like accuracy and latency, highlighting the comprehensive improvement potential of our novel STR technique in the SNN landscape. 
{
    \small
    \bibliographystyle{ieeenat_fullname}
    \bibliography{main}
}

\end{document}